\pdfoutput=1
\documentclass[11pt]{article}
\usepackage{setspace}
\usepackage{sectsty}
\usepackage{amsfonts,amsmath,amssymb,url,epsfig,float}
\usepackage{geometry} 
\usepackage[normalem]{ulem}
\usepackage[table]{xcolor}
\usepackage{verbatim}
\usepackage{hyperref}
\usepackage{soul}
\usepackage{bbm}
\usepackage{booktabs}
\usepackage{tabularx}
\usepackage{cleveref}
\usepackage[table]{xcolor}

\sectionfont{\sffamily\bfseries\upshape\large}
\subsectionfont{\sffamily\bfseries\upshape\normalsize} 
\subsubsectionfont{\sffamily\mdseries\upshape\normalsize}
\makeatletter
\renewcommand\@seccntformat[1]{\csname the#1\endcsname.\quad}
\makeatother

\usepackage{natbib}

\makeatletter
\def\@maketitle{%
  \begin{center}%
  \let \footnote \thanks
    {\large \@title \par}%
    {\normalsize
      \begin{tabular}[t]{c}%
        \@author
      \end{tabular}\par}%
    {\small \@date}%
  \end{center}%
}
\makeatother

\newcommand{\btR}{\vspace{-.25in}\begin{quotation}\begin{small}\noindent\begin{verbatim}}

\title{\bf Conformal Prediction and Human Decision Making\footnote{We thank John Cherian, Tiffany Ding, and Jason Hartline for feedback on a draft.}\vspace{.1in}}

\author{Jessica Hullman, Yifan Wu, Dawei Xie, Ziyang Guo, Andrew Gelman \vspace{.1in}}

\date{7 Mar 2025\vspace{-.2in}}

\newcommand{\rpos}{\textrm{R}}
\newcommand{\rprior}{\mathrm{R}_{\varnothing}}

\newcommand{\mstate}{s}
\newcommand{\statespace}{\mathcal{S}}
\newcommand{\mstateRV}{S}

\newcommand{\joint}{p}

\newcommand{\dist}{p}

\newcommand{\distover}[1]{P(#1)}

\newcommand{\action}{a}
\newcommand{\actionspace}{\mathcal{A}}

\newcommand{\signal}{v}
\newcommand{\signalspace}{\mathcal{V}}
\newcommand{\signalRV}{V}

\newcommand{\data}{x}
\newcommand{\dataspace}{\mathcal{X}}
\newcommand{\xb}{\textbf{x}}
\newcommand{\xnew}{X_{\text {new}}}
\newcommand{\dataRV}{X}

\newcommand{\ystate}{y}
\newcommand{\ystatespace}{\mathcal{Y}}
\newcommand{\ypredict}{\hat{y}}
\newcommand{\ynew}{Y_{\text {new}}}
\newcommand{\ystateRV}{Y}

\newcommand{\hdataspace}{\mathcal{W}}
\newcommand{\wb}{\textbf{w}}
\newcommand{\hdataRV}{W}

\newcommand{\score}{L}

\newcommand{\infoval}{\Delta}

\DeclareMathOperator*{\argmin}{arg\,min}

\newcommand{\dtr}{\mathcal{D}_{\text{tr}}}
\newcommand{\dcal}{\mathcal{D}_{\text{cal}}}

\newcommand{\model}{\hat{f}}
\newcommand{\modelprob}{\hat{p}}

\newcommand{\cset}{\hat{\mathcal{C}}}

\newcommand{\test}{T}
\newcommand{\tests}{\mathcal{T}}

\usepackage{ifthen}

\newcommand{\expect}{\mathbb{E}}

\newcommand{\prob}[2][]{\text{\bf Pr}\ifthenelse{\not\equal{}{#1}}{_{#1}}{}\![{\def\givenn{\middle|}#2}]}

\DeclareMathOperator{\quantile}{quantile}

\newcommand{\compatiblejoint}{\mathcal{P}}

\begin{document}\sloppy
\maketitle

\begin{abstract}
Methods to quantify uncertainty in predictions from arbitrary models are in demand in high-stakes domains like medicine and finance. 
Conformal prediction has emerged as a popular method for producing a set of predictions with specified average coverage, in place of a single prediction and confidence value. 
However, the value of conformal prediction sets to assist human decisions remains elusive due to the murky relationship between coverage guarantees and decision makers' goals and strategies. How should we think about conformal prediction sets as a form of decision support? 
We outline a decision theoretic framework for evaluating predictive uncertainty as informative signals, then contrast what can be said within this framework about idealized use of calibrated probabilities versus conformal prediction sets. Informed by prior empirical results and theories of human decisions under uncertainty, we formalize a set of possible strategies by which a decision maker might use a prediction set. 
We identify ways in which conformal prediction sets and posthoc predictive uncertainty quantification more broadly are in tension with common goals and needs in human-AI decision making. We give recommendations for future research in predictive uncertainty quantification to support human decision makers. 

\end{abstract}

\section{Quantifying uncertainty in black-box predictions}
What does it mean to target a quantification of prediction uncertainty to a decision maker's task? Can we do this using general methods that work regardless of the specific decision setting? These statistical questions are as relevant as ever as machine learning models are deployed to assist human decision makers in domains like medicine, criminal justice, and finance. Methods for rigorously quantifying uncertainty in model predictions---even if the model is a black box---are in high demand.

Consider for example a telehealth doctor who can access predictions from a computer vision model while evaluating images of a skin condition submitted by a prospective patient. Presenting information about the model's confidence in its predictions is generally thought to help the human expert use the information more effectively. If, for example, the model assigns a moderate probability to a relatively rare but deadly condition, the doctor might advise them to schedule an appointment directly with a specialist rather than first seeing a primary care physician. 
Presenting uncertainty may also motivate ``practico-epistemic'' actions performed for the sake of epistemic gain, such as when the ambiguity about the true condition motivates the doctor to gather more evidence, as well as useful diversification in the choice of action~\citep{manski2009}, such that the doctor varies their advice across similar patients to balance expected false positives and false negatives in the face of ambiguity about the best treatment policy.  

A common minimum goal for such settings is that a
predictive model is \textit{calibrated}: the rate at which events predicted by the model are realized approximately matches the probability assigned by the model to those events~\cite{dawid1985calibration,foster1998asymptotic}. 
Here we focus on multiclass prediction, where we assume a classification model $\model$ trained on historical data. When we apply the model to an input feature vector $\xb = (x_1,x_2,...,x_d) \in \dataspace \subseteq \mathbb{R}^d$, we obtain a predicted label $\model(\xb)=\ypredict \in \ystatespace$, but also have access to a predicted probability $\modelprob_{y}(\xb)=Pr(\ystateRV=\ystate|\xb)$ for each possible label $\ystate \in \ystatespace$. Calibration in this setting means that for all predicted values $b \in [0,1]$, for all pairs of inputs and labels $(\dataRV=\xb, \ystateRV=\ystate)$ for which $\modelprob_y(\xb)=b$, then $(b\times100)\%$ of the time, $\ystate$ is the true label.
Calibration is useful when we expect decisions to be based only on predictions, because even if decision makers weigh errors differently than the machine learning model, calibrated predictions allow them to trade off their true risk according to their own decision problem.

However, some machine learning paradigms result in predictive uncertainty estimates that are not calibrated.  
Deep learning models for multi-class classification, for example, produce pseudo-probabilities with predictions called softmax values. 
These are achieved by normalizing the activations produced by the last hidden layer in the neural network to a $[0,1]$ range, 
but this normalization does not enforce calibration with reality or even internal consistency:  these numbers when will not in general follow the laws of probability (that $\mbox{Pr}(A \vee B)= \mbox{Pr}(A) + \mbox{Pr}(B)$ if the events $A$ and $B$ are disjoint, $\mbox{Pr}(A \wedge B)= \mbox{Pr}(A)\times\mbox{Pr}(B)$ if they are independent). 
Beyond their lack of coherence, insufficient data, dataset biases, or implications of particular choices of architectures, hyperparameters, or model classes can make model confidence estimates vulnerable to overfitting, and thus unreliable for decision making~\citep{guo2017calibration}. 
Bayesian methods like Bayesian neural networks~\citep{neal2012bayesian, gal2016dropout} can provide more meaningful probabilities by approximating a distribution over network weights. 
However, their computational requirements in high-dimensional settings like images or text limit their use. 

In theory, empirically calibrated probabilities can be also achieved by training models with a proper loss function~\citep{blasiok2024does}, or applying calibration procedures~\citep{hebert2018multicalibration,guo2017calibration,gopalan2022low} that learn adjustments to model-predicted probabilities to achieve calibration on held-out data. 
For example, recent work in theoretical computer science demonstrates the possibility of learning multicalibrated predictors--predictors that are simultaneously calibrated for all intersecting groupings of the data, not just marginally~\citep{hebert2018multicalibration}--by iteratively post-processing predictions via a boosting procedure.  
However, the amount of data required to achieve such guarantees renders them impractical in common scenarios like multiclass classification with a moderately large label space. 





\begin{figure}[t]
  \centering
  \includegraphics[width=1.0\linewidth]{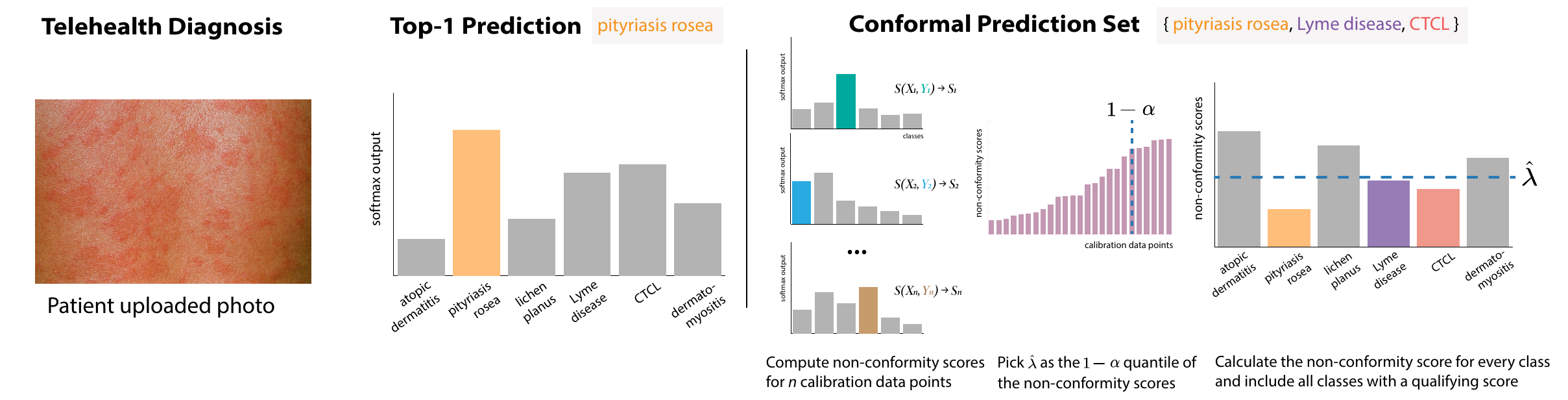}
    \caption{Conventional (top-1) prediction (left) compared to split conformal prediction (right) using a nonconformity score based on softmax output.}
    \label{fig:fig1}
\end{figure}

Enter conformal prediction, a class of distribution-agnostic approaches for quantifying uncertainty in predictions that produces \textit{sets} of predictions that promise to contain the true label with a user-defined probability $1-\alpha$. In contrast to returning a single top prediction with the highest model confidence score (Figure~\ref{fig:fig1} left), split conformal prediction algorithms return a set of predictions designed to achieve $1-\alpha$ coverage on average (Figure~\ref{fig:fig1} right). 
In the common variant of split conformal prediction with marginal coverage, this set is constructed by comparing a score, representing the ``non-conformity" of the new instance with what the model has learned, to a threshold chosen to achieve the expected coverage over the randomness in the selection of test points and a set of held-out calibration instances for which ground truth is known.

Conformal prediction has been argued to provide meaningful uncertainty quantification to improve decisions from predictions in scenarios where models are deployed to assist doctors or other experts \citep{banerji2023clinical,kiyani2025decision,lu2022fair, straitouri2023improving}. 
Relative to a single top prediction with a confidence value, prediction sets may better convey to decision makers that multiple labels are consistent with the available information. By providing a finite set of possibilities with some probabilistic assurance, they can guide decision makers to focus their attention on a limited number of possibilities. For example, in the absence of a prediction set, which possibilities the telehealth doctor considers in the limited time they are given to assess each patient might depend on the vagaries of the cases they have seen in the recent past.       
Despite the difficulty of properly interpreting coverage guarantees based on long-run frequencies (as has been observed even with researchers for $95\%$ confidence intervals~\cite{belia2005researchers,hoekstra2014robust}, doctors and other experts may at least find the notion of $1 - \alpha$ coverage familiar for signaling uncertainty.

At the same time, a clear account of how exactly conformal prediction sets may impact human decisions relative to alternative presentations of predictive uncertainty remains lacking.  
While a calibrated probability has a straightforward translation as a decision aid, 
a prediction set offers no measure of the credibility of items within the set, leaving it in the hands of the decision maker to decide a strategy for choosing between them. Recent work observes that prediction sets are optimal for conservative decision makers who seek to act in a way that maximizes their worst case utility~\citep{kiyani2025decision}, but this strategy may or may not align with needs in a particular use case. 
We ask several broader questions with an eye toward how experts' domain knowledge may interact with prediction set design: What might an expert's decision processes with a conformal prediction set entail? 
How should conformal prediction algorithms be designed in light of common expectations that human experts will bring unique knowledge beyond that available to an AI model?   
We outline a decision theoretic framework for evaluating predictive uncertainty as informative
signals, then contrast what can be said within this framework about idealized use of calibrated
probabilities versus conformal prediction sets. 
Informed by prior empirical results and theories
of human decisions under uncertainty, we highlight how optimal use of a conformal prediction set by a decision maker is underdefined, and present possible strategies by which a decision maker might use a prediction set. 
Taking a step back, we consider what kinds of philosophical commitments conformal prediction and other post hoc calibration approaches entail, and reflect on how theoretical work on black-box uncertainty quantification can neglect to consider important practical features of an approach.
We synthesize future directions for research aimed at better aligning conformal prediction sets with the needs of human decision makers.


This paper is organized as follows. First, we briefly summarize common variants of conformal prediction with an emphasis on split conformal prediction (Section~\ref{sec:background}). 
We introduce a decision-theoretic framework for evaluating decisions under uncertainty (Section~\ref{sec:evaluation}), then consider the value of a calibrated prediction to a decision maker (Section~\ref{sec:calibration}), contrasting the case where they do and do not have access to information external to the model predictions. 
We consider several possible models of decision making with conformal prediction sets (Section~\ref{sec:conformal}).
We conclude by discussing future directions for research on human-aligned predictive uncertainty (Section~\ref{sec:discussion}).

\section{Split conformal prediction}
\label{sec:background}
Introduced by \citet{vovk2005algorithmic} as a distribution-free statistical framework for quantifying prediction uncertainty, conformal prediction is a model-agnostic approach to generating prediction sets (or intervals) expected to contain the true label (or real-valued response) with a user-defined probability. 
Formally, to predict the label $\ynew\in\ystatespace$ of a new instance $\xnew\in\mathcal{X}$ with some classifier $\model$ (where capital letters are used to denote the random variables), conformal prediction returns a set $\cset(\xnew) \subseteq\mathcal{Y}$ such that \textit{on average} 
the following coverage guarantee holds: 
\begin{equation}
\label{eq:marginal}
\mbox{Pr}\left(\ynew \in\cset\left(\xnew\right)\right) \geq 1-\alpha
\end{equation}
where $\alpha$ is the user-specified error rate. This \textit{marginal coverage} guarantee is expected to hold over the randomness in the selection of test points and data used by the algorithm to achieve the guarantee.
While coverage could be achieved trivially by outputting the entire label space as a prediction set, ideally the set size is not larger than necessary and is adaptive in the sense that larger sets signal higher predictive uncertainty.

Split (a.k.a., \textit{inductive}) conformal prediction~\citep{vovk2005algorithmic, angelopoulos2023conformal} for classification, which we focus on here, 
uses a held-out calibration set to post-process model predictions, generating sets expected to achieve marginal coverage at the target level.
Given a classifier $\model$ trained on a training dataset $\dtr$, we identify a calibration dataset of the form $\dcal=\{(X_i,Y_i)\}_{i=1}^{n}$ that is disjoint with $\dtr$ (i.e., $\dtr \cap \dcal = \emptyset$). No restriction is placed on the relationship between $\dtr$ and $\dcal$; they may be generated by different distributions. 

Next, we select a heuristic of predictive uncertainty that can be obtained from $\model$.  
For example, we might use $\modelprob_{y_i}(\xb) = \sigma(\textbf{z})_i$, the softmax probability for classifying $\xb$ as $y_i$, where $i$ also indexes the corresponding class output $z_i$ in the model's last layer outputs $\textbf{z}$\footnote{Given a vector of unnormalized log probabilities $\textbf{z} = [z_1, ..., z_k$] representing outputs of the neural network's last layer for an input $\xb$ and classes $[y_1, ..., y_k]$, the softmax probability $\sigma(\textbf{z})_i$ for a class indexed by $i$ is $\frac{e^{\textbf{z}_i}}{\sum_{j=1}^{k}e^{\textbf{z}_j}}$.}. Typically, we assume $\modelprob(\xb)$ to be the vector of class probabilities for $\xb$ after temperature scaling (a.k.a. Platt scaling)~\citep{platt1999probabilistic,guo2017calibration}.   
This heuristic measure of uncertainty is transformed to create a 
\textit{non-conformity score} function $S: \dataspace \times \ystatespace \rightarrow \mathbb{R}$ that can be applied to any pairing of instance features $\xb$ and true label $y$; for example, $S(\xb, y) = 1-\modelprob_{\ystate}(\xb)$.

We then estimate the $1-\alpha$ quantile of the $n$ non-conformity scores $S_1, \ldots, S_n$ and their corresponding labels from $\dcal$. 
Specifically, a threshold $\hat{\lambda}$ is set to be the $\lceil(1-\alpha)(n+1)\rceil / n$ quantile\footnote{$\hat{\lambda}$ is set to be the $\lceil(1-\alpha)(n+1)\rceil / n$ quantile when $\alpha \geq 1/(n+1)$; otherwise $\hat{\lambda}$ is set to $+\infty$.} to correct the threshold for the finite sample. The prediction set for $\xnew$ can then be constructed by including all labels with a non-conformity score less than or equal to $\hat{\lambda}$:
\begin{equation}
\label{eq:cpset}
\cset\left(\xnew\right)=\left\{y\in\ystatespace: S\left(\xnew, y\right) \leq \hat{\lambda}\right\} 
\end{equation}

The key assumption behind split conformal prediction's coverage promise is that the 
data in $\dcal\cup \left\{\left(\xnew, \ynew\right)\right\}$ are statistically \textit{exchangeable}, such that permuting the order of these data points leaves their joint distribution unchanged. 

Since the data points in $\dcal\cup \left\{\left(\xnew, \ynew\right)\right\}$ are exchangeable, the same conditions hold for their non-conformity scores $\{S_i\}_{i=1}^n\cup \{S_{\text {new}}\}$. To get an intuition for why this is sufficient to achieve coverage, consider what it means for random variables $S_1,S_2,...,S_{n+1}$ to be exchangeable  (where $S_{n+1}$ represents $S_{\text {new}}$). 
Exchangeability means that the rank of the random variable $S_{n+1}$ is equally likely to take any value from $1,...,n+1$. This implies that the probability that $S_{n+1}$ is less than or equal to the $\lceil(1-\alpha)(n+1)\rceil$ smallest score of $\{S_i\}_{i=1}^n\cup \{S_{n+1}\}$ is greater than or equal to $1-\alpha$. Further, if the previous holds this means that $S_{n+1}$ is less than or equal to the $\lceil(1-\alpha)(n+1)\rceil$ smallest score of $\{S_i\}_{i=1}^n$. Consequently, the probability that $S_{n+1}$ is less than or equal to the $\lceil(1-\alpha)(n+1)\rceil$ smallest of $S_1,S_2,...,S_{n}$ is greater than or equal to $1-\alpha$ over repeated sampling of the calibration and test instances. 
Before proceeding, we make a few remarks about key components of conformal prediction.

\vspace{3mm}
\noindent\textbf{Score function.} No formal restrictions are put on the score function used to construct sets, yet the score function should empirically capture the uncertainty around model predictions such that larger scores signal higher uncertainty. However, the usefulness of the prediction sets will naturally vary with the informativeness of the score function. 
For neural networks, the score function is commonly derived from the model outputted softmax scores. 
The simplest such score function is simply $1-\modelprob_{\ystate}(\xb)$. 
Alternatively, to improve the adaptability of the sets--the relationship between set size and the difficulty of the instance--one can use score functions that use the cumulative softmax probabilities ranked in descending order, i.e., from the most likely class to the least likely class, until the true label is reached~\citep{romano2020classification, angelopoulos2021uncertainty}.


\vspace{3mm}
\noindent\textbf{Exchangeability.} 
An obvious concern in practice is that exchangeability may be violated, such as when the covariate, label, or posterior distribution shifts over time. This possibility inspires conformal methods that learn weighting schemes by leveraging additional knowledge about data shifts
~\citep{tibshirani2019conformal,podkopaev2021distribution,barber2023conformal,prinster2024conformal}. 
A distinct line of work focuses on online conformal prediction algorithms~\citep{gibbs2021adaptive,feldman2023achieving,bhatnagar2023improved,gibbs2024conformal,angelopoulos2024online} that adapt regret minimization from online learning to employ adaptive thresholds that are adjusted upon observing the miscoverage of recent instances. 

\vspace{3mm}
\noindent\textbf{Marginal coverage and beyond.} 
The marginal coverage guarantee associated with split conformal prediction provides a sort of compromise in light of the difficulty of achieving calibrated prediction probabilities conditional on groupings that can be defined on the feature space $\mathcal{X}$.  
The expected $1-\alpha$ coverage probability of vanilla split conformal prediction is over the random selection of calibration instances and test points. 
This is a weaker condition than coverage conditioned on a specific instance:  
 \begin{equation}
 \label{eq:conditional}
 \mbox{Pr}\left(\ynew \in \cset\left(\xnew\right) \mid \xnew=\xb\right) \geq 1-\alpha, \text { for any distribution, for almost all } \xb
 \end{equation}

Conditional coverage is not possible without making distributional assumptions~\citep{vovk2012conditional,barber2021limits}. 
However, decision makers may naturally desire  instance-specific uncertainty, and in the applied literature, split conformal prediction is occasionally misinterpreted as providing ``personalized" or ``individualized" measures of uncertainty (e.g.,~\citep{banerji2023clinical,garcia2023uncertainty}). 
However, the coverage of split conformal prediction actually describes what we expect on unlimited test instances 
given \textit{repeated resampling} of the calibration set from the same distribution over $\dataspace \times \ystatespace$. 
The deviation from expected coverage in any particular application conditional on $\dcal$ shrinks with rate $n^{-1/2}$ as the size of $\dcal$ increases. Marginal coverage also means that the attained coverage can vary considerably over regions of the feature space, such that some combinations of features are overcovered and others undercovered.
Both facts are overlooked when the $1-\alpha$ coverage guarantee is mistakenly described as applying to the specific prediction set.  
These errors parallel misinterpretations of the stated confidence level (e.g., 95\%) of Frequentist confidence intervals as applying to the specific interval at hand~\citep{hoekstra2014robust}.

To alleviate practical limitations of marginal coverage, multiple variations on conformal prediction~\citep{barber2021limits,jung2023batch,gibbs2023conformal} pursue an approximate version of conditional coverage where the coverage guarantee is conditional on the new instance's group membership, rather than the instance itself, i.e.,
\begin{equation}
\label{eq:groupconditional}
\mbox{Pr}\left(\ynew \in \cset\left(\xnew\right) \mid \xnew \in G\right)=1-\alpha, \; \forall G \in \mathcal{G}
\end{equation}
where $\mathcal{G}\subseteq 2^{\mathcal{X}}$ is some collection of pre-specified groups. As in the case of multicalibration algorithms, this stronger form of coverage requires more data to achieve. 

\section{Preliminaries: Evaluating decision making under uncertainty}
\label{sec:evaluation}

To justify methods that produce quantified prediction uncertainty for decision making requires a way to evaluate the quality of decisions, and to appraise the value of specific information for a decision task. What characterizes a good measure of decision quality?

We can get some intuition for what is needed by considering how simple approaches fall short. Consider the use of post hoc decision accuracy, a common approach in research on AI-assisted decision-making. This approach measures the accuracy of a human decision-maker's final decisions given access to instance information as well as the human's own prediction and the prediction of a model for each instance. For example, in the case of the telehealth doctor making a diagnosis decision given a patient's uploaded photo of their condition, one might look at the proportion of cases in some labeled set of decision trials where the doctor chose the correct diagnosis after viewing the instance and model prediction. 

There are two reasons post hoc accuracy is problematic as a general measure of decision quality.  
First, evaluating a decision against the realized outcome is problematic whenever outcomes are probabilistic.  
To see this, consider how the doctor might correctly observe that conditional on features present in the the image, melanoma is the most likely diagnosis (i.e., $\mbox{Pr}(\ystate = \mbox{melanoma}|\xb) > \mbox{Pr}(\ystate = \ystate_0|\xb), \forall \ystate_0 \in \ystatespace \text{ s.t. } \ystate_0 \neq \mbox{melanoma}$). 
However, it could be the case that while the observed features most often co-present with melanoma, they are also associated with psoriasis, and psoriasis happens to be the true diagnosis for this patient. 
Saying that the doctor is wrong for acting as if melanoma is the diagnosis based on decision accuracy in hindsight fails to consider that the beliefs they arrived at were rational---that is, they captured the probability of each diagnosis under the true data-generating process---given the information available at the time.

Second, using post hoc accuracy to measure decision quality does not allow for possibly different costs of different types of errors in a decision scenario. If the cost of not testing for melanoma is high (because valuable time is lost in the event that it is the true diagnosis), then even when psoriasis is the more likely condition a priori, a doctor may still be better off following up as if melanoma were the diagnosis.

Statistical decision theory \citep{wald1949statistical,savage1972foundations} provides a framework for evaluating decisions that accounts for both of these considerations. Below we define a decision problem and show how a definition of normative decision-making can be used to assess the value of predictive uncertainty information. 


\subsection{Definition of a decision problem}
We assume a decision maker who seeks the action that maximizes their expected reward under a given utility function and data-generating process. Decisions are evaluated based on their expected utility, accounting for uncertainty about the outcome.

Formally, a decision problem assumes a \textit{state} of the world, modeled as a random variable $\mstateRV$, whose realized value is from a discrete set of possible (mutually-exclusive) states ($\mstate \in \statespace$). The realized state is unknown at the time of the decision, but follows a distribution $\joint(\mstateRV)$.  
The decision maker faces a choice of \textit{action} from a set of possible actions, $a \in \actionspace$. Decision quality is assigned by a loss function that assigns a numeric score representing our loss from making a given decision under a given (realized) state of the world: $\score: \actionspace \times \statespace \rightarrow \mathbb{R}$.

This abstraction provides sufficient flexibility to model a variety of decision problems. We can incorporate actions that involve epistemically-motivated behaviors like collecting additional data by including abstention in the action space. For example, in the example of medical diagnosis, we might imagine a two stage decision problem where the first action set includes an option to abstain from diagnosing in order to gather more information, which carries some cost that increases the ultimate decision loss.  

Next we consider the structure to the information that is available to the decision maker at decision time (such as the instance features, a model prediction, and a quantification of the prediction uncertainty). Specifically, we define the \textit{data-generating model} $\joint$, the joint distribution of the signals and state over $\signalspace \times \statespace$ from which the distribution of the state $\joint(\mstateRV)$ is derived.  
This joint distribution assigns to each possible pairing of signal $\signalRV=\signal$ and state $\mstateRV=\mstate$ a probability $\joint(\signal,\mstate)$.
Slightly abusing the notation, we use $\joint(\signalRV)$ to denote the marginal distribution of $\signalRV$ with expectation over $\mstateRV$ and $\joint(\mstateRV)$ to denote the marginal distribution of $\mstateRV$ with expectation over $\signalRV$.
We can thus use $\joint(\mstate)$ and $\joint(\signal)$ to represent the marginal probabilities of $\mstateRV=\mstate$ and $\signalRV=\signal$ respectively.

\subsubsection{Decision rule}
Given a decision problem, a decision strategy or rule is a function that maps from a signal representing the information available to the decision-maker to a specific choice of action. The space of all possible decision strategies is intractable to reason about. Instead, researchers are often interested in either identifying what kinds of simple constraints characterize normative decision-making, which we define below, or in empirically describing decision rules that human decision-makers employ.



\subsection{Using normative decision-making to quantify the value of predictive uncertainty}
The joint distribution $\joint$ captures the information structure of the decision problem. We can identify the best possible performance of any decision maker on a decision problem by considering how well we expect an idealized agent who knows the true data-generating model to do. 
This is captured by the expected loss of a Bayesian rational decision maker who has knowledge of the data-generating model and access to the signal for each particular instance. 
This decision maker starts with knowledge of the prior over states $\joint(\mstateRV)$, processes the signals to extract the relevant information about the uncertain state, and updates their prior to arrive at posterior beliefs for each state $\mstate \in \statespace$: 

\begin{equation}
\joint(\mstate|\signal) = \frac{\joint(\signal,\mstate)}{\sum\nolimits_{\mstate \in \statespace} \joint(\signal, \mstate)}
\label{eq:posterior}
\end{equation}

\noindent They select the loss-minimizing action under those beliefs: 

\begin{equation}
\action_{opt}(\signal) = \argmin_{\action\in\actionspace} \expect_{\mstateRV\sim\joint(\cdot|\signal)}({\score(\action,\mstateRV)})
\label{eq:opt}
\end{equation}

\noindent where $\joint(\cdot|\signal)$ is the posterior over states. These assumptions can be used to derive performance benchmarks defined over a decision problem~\citep{wu2023rational}. 
The best attainable performance of any decision maker facing the decision problem is upper bounded by the expected loss of the rational Bayesian agent with access to the signal: 

\begin{equation}
\rpos(\joint, \score)
= \expect_{\signalRV, \mstateRV \sim \joint}(\score({\action_{opt}(\signalRV),\mstateRV)) }
\label{eq:benchmark}
\end{equation}

Variations on this approach relax the assumption that the decision maker knows the DGM, and instead calculate the expected performance of a rational decision maker given some approximation $\hat{p}$ of $p$ based on the available information (such as a set of labeled examples that the decision maker has previously seen)~\cite{hullman2024decision}.

We can use this Bayes rational benchmark to evaluate different forms of uncertainty information we might present to decision makers ex ante. 
We quantify the value of information $\infoval$ presented by signal $\signalRV$ as the difference in the expected loss of the rational decision maker when they have access to that information (as captured by Equation~\ref{eq:benchmark}) versus when they do not:

\begin{equation}
    \infoval(\joint, \score) =  \rprior(\joint, \score) - \rpos(\joint, \score)
\end{equation}

\noindent where the rational baseline $\rprior$~\citep{wu2023rational} is the expected loss of the rational Bayesian decision maker who must choose the best fixed action under knowledge of only the prior $\joint(\mstateRV)$:

\begin{equation}
\rprior(\joint, \score)
=  \min_{\action\in \actionspace}\expect_{\mstateRV\sim \joint(\cdot)}[\score(\action, \mstateRV)]
\label{eq:baseline}
\end{equation}

 The value of information $\infoval$ quantifies the ``room'' for reduction in expected loss associated with a particular type of signal. We can get a sense of the ex ante maximum possible ``relevance'' of different forms of prediction uncertainty by looking at how the expected loss of the idealized decision maker changes as we change the nature of the predictive uncertainty included in the signal. 

Below, we use the definition of a decision problem, decision rule, and rational decision-maker to contrast decision strategies for calibrated probabilities to decision strategies for prediction sets. We consider  perspectives based in normative models as well as by empirical descriptions of human decision-making under uncertainty.

\section{Decisions from calibrated probabilities}
\label{sec:calibration}
Note that the lower bound on the expected loss of an idealized rational decision maker given some signaling strategy $\signalRV$ (Equation~\ref{eq:benchmark}) assumes that the decision maker arrives at posterior beliefs (Equation~\ref{eq:posterior}) using their knowledge of the joint distribution $\joint$. However, decision makers in the world will not generally possess perfect information about the data-generating model. Consider a Bayes rational decision maker who must decide given only the signal for each instance, without knowing the data-generating model. What is the ideal form of uncertainty information to provide such a decision maker?

From the above formulation of idealized decision making, it follows that the posterior probability distribution over states given the signal---$\joint(\mstateRV|\signal)$---is sufficient information for such a decision maker to optimize. This is because when we expect the decision maker to choose the action with the highest expected reward conditional on these beliefs, this information is equivalent to telling them the action that maximizes their payoff. 
Hence, among signals conveying probabilistic information about the state, the posterior probability of the state is maximally informative. 

This fact has implications for the value of calibrated predictions for decision making. Imagine that for each individual case we will predict on, there is a true probability distribution over outcomes, from which the realized outcome is drawn. 
When predicted probabilities are calibrated, they are ``outcome 
indistinguishable''~\citep{dwork2021outcome} from the true distribution, meaning that even given full access to the predicted model and the historical outcomes, 
it is not possible to invalidate the model's predictions.\footnote{More precisely, a predictor $\modelprob$ is outcome indistinguishable with respect to a family of tests $\tests$ if samples from the true distribution $(i, \ystate_i) \sim  D^*$ cannot be distinguished by $\tests$ from samples from the predictor's distribution $(i, \ypredict_i) \sim  D(\modelprob)$. In other words, for each test $\test \in \tests$, the probability that $\test$ outputs 1 is (nearly) the same on samples
from $D^*$ and $D(\modelprob)$.} 
For this reason, computer science theorists associate outcome indistinguishability with trustworthiness of predictions for decision making (e.g.,~\citep{gopalan2021omnipredictors,hu2024predict,zhao2021calibrating}). 


But the ex ante value that we expect outcome indistinguishability to afford calibrated predictions for decision makers is not without assumptions. 
We must distinguish between two types of settings: those where the decision maker does not have additional knowledge over that available to the model, and those where they do. Only in the former does a calibrated probability directly ensure trustworthiness for decision making.

To ground the discussion, consider the example setting of the telehealth doctor facing a choice of diagnosis, where the uncertain state $\mstate \in \statespace$ is the true label $y$ from label space $\ystatespace$ and the action space $\actionspace$ is $\ystatespace$. They make their choice in light of information about the specific instance $\xb \in \dataspace$, associated with true label $\ystate$ generated by some true process $\joint \in \distover{\dataspace \times \ystatespace}$.  We might expect that based on their domain experience, the human has some internal approximating model of the data-generating process which enables them to make a prediction $\ystate^{H}$ about the label. They also have access to the prediction $\ystate^{AI}$ output by a model (or AI) trained on what is deemed to be relevant prior information.

In the first setting, the decision maker does not have additional knowledge over that available to the model. In other words, they are restricted to having at most equivalent decision-relevant information to the predictive model. 
Both the human prediction and the AI prediction are generated from a common set of features: $(\dataRV, \ystateRV^{H}) \sim \joint^{H}$ and $(\dataRV, \ystateRV^{AI}) \sim \joint^{AI}$, where $\joint^H, \joint^{AI} \in \distover{\dataspace \times \ystatespace}$. Or, the human may only have access to some subset of the features that the model accesses, such as when the feature space is high dimensional as in classifying images or text, i.e., 
$\joint^H \in \distover{\dataspace^H \times \ystatespace}$ where $\dataspace^H \subseteq \dataspace$.
This setting has been well studied under the guise of clinical versus statistical prediction~\citep{meehl1954clinical}, where a number of empirical studies have found human decisions tend to underperform those of statistical models on average when the test instances are drawn from the same distribution~\citep{aegisdottir2006meta,grove2000clinical}. 
This finding should not come as a surprise: when statistical models are evaluated on the same distribution on which they are trained to predict, and the human possesses no additional information, statistical prediction via loss minimization defines the standard for efficient use of the information.

In settings where we are confident that these conditions hold, we should expect the best decision quality from automation. However, full automation may not be an option for reasons of liability. The healthcare organization may not be comfortable, for example, making telehealth referral decisions about patients' conditions without a human doctor in the loop. Consequently, the goal becomes how to help the human realize that it is in their best interest to trust the model predictions. Ideally, if we provide them with calibrated probabilities, they will use these directly to optimize.


In addition to being critical to arguments that calibration makes predictions trustworthy for decision making, the assumption that the decision maker does \textit{not} have access to relevant additional information also implicitly motivates broader arguments that providing model-generated uncertainty information to users of predictive models supports appropriate trust~\citep{bhatt2021uncertainty}. As in the explainable AI literature, however, this assumption is rarely made explicit.

\subsection{Decisions under calibrated probabilities plus private signals}
A challenge with the assumptions of a decision maker who relies only on the predictions of a model is that often in practice, we \textit{want} the decision maker to consider other signals as well. 
What happens if a human expert is expected to have access to additional information beyond that available to train the model? 
In many scenarios where experts might be helped by predictive models, such as medicine or criminal justice, there is a latent belief that the expert \textit{will} have access to further information. A doctor or judge, for example, may interact with a patient or defendant to gather information that is not available in their record. 
It is often speculated that they may observe signals---like the patient's attitude or the defendant's expressions of remorse---that contain decision-relevant information, but which are hard to measure or formalize~\citep{meehl1954clinical,collina2024tractable,guo2024decision,steyvers2022bayesian,wilder2020learning}. Until very recently, few attempts have been made to provide organizations with sufficient methods to identify when human decision makers appear to have additional information that could lead to complementary performance with an AI~\citep{alur2024auditing,guo2024decision,guo2024unexploited}.
Hence it is often difficult to rule out this setting in practice.

Consider a decision maker with access to some auxiliary private information (i.e., set of features) $\wb \in \hdataspace \subseteq \mathbb{R}^d$ for each instance, where $\hdataspace \cap \dataspace = \emptyset$. Hence, the decision maker has access to instance information $\dataRV^H = \{\dataRV, \hdataRV\}$, and their predictions $(\dataRV^H,\ystateRV^H) \sim \joint^H$ with $\joint^H \in \distover{\dataspace^H\times  \ystatespace}$ and associated posterior distribution $\joint^H(\ystateRV|\dataRV^H)$ are generated from this full set of information.
Imagine that at decision time, given instance information $\xb$ they have access to a display that also provides them with the model's top predicted class and confidence: $\signal = \{\xb, \ypredict, \joint^{AI}(\ypredict|\xb)\}$. For simplicity, imagine a binary diagnosis problem such that $\ystatespace = \{0,1\}$ such that $\joint^{AI}(\ystate|\xb)$ is the predicted probability that the patient has the disease given instance $\xb$. Assume $\joint^{AI}(\ystateRV|\dataRV)$ is calibrated on data-generating process $\joint$. 
If $\joint^{AI}(\ystate|\xb)$ was the same as the probability conditional on all of the available information 
($\joint^{AI}(\ystate|\xb^H) = \joint^{AI}(\ystate|\xb,\wb))$, the utility-maximizing decision would correspond to placing a threshold $j^* \in [0,1]$ on the predicted probability, representing the point at which the expected value of choosing $\action=1$ is equivalent to that of $\action=0$. 

However, because the model's predictions are only calibrated over $\dataRV$, not necessarily over $\dataRV^H$, this will not necessarily be the case, and the decision maker can end up making a worse decision by trusting the probability over their own information. For example, imagine that the decision maker's private information $\hdataRV$ correlates more strongly with the state than does $\dataRV$. Then we would expect thresholding the model's predicted probability to result in lost utility. The benchmark in Equation~\ref{eq:benchmark} corresponding to signals taking the form $\signal=\{\xb,\ypredict,\joint^H(\ystate|\xb^H)\}$ will be greater than or equal to that corresponding to signals taking the form $\signal=\{\xb,\ypredict,\joint^{AI}(\ystate|\xb)\}$. The biggest possible loss we could incur from showing the model predicted probability rather than the correct probability corresponds to the case where the private information is perfectly revealing of the state ($\hdataRV = \ystateRV$) whereas the unidimensional feature $\dataRV$ is uninformative. 

Can this possibility be avoided? When the decision maker has access to private signals in addition to the signals available to the prediction model, the maximally informative signal for a rational decision maker combines the auxiliary private information with the information provided by an AI prediction. In other words, a maximally informative predictor is multi-calibrated with respect to both the human's private information $\hdataRV$ and the features $\dataRV$ that both the human and the AI can access. In the case where the information in the feature space is hard to extract, e.g.\ with high-dimensional $\hdataspace$ and $\dataspace$ such as image data, achieving multi-calibration with respect to both the human's  predictor $\joint^H(\ystateRV|\dataRV^H)$ and the AI's predictor $\joint^{AI}(\ystateRV|\dataRV)$ may be more realistic than multi-calibration over $\hdataRV$
and $\dataRV$. A predictor $\joint$ is multi-calibrated with respect to predictors $\joint^H$ and $\joint^{AI}$ if it is conditionally correct:
\begin{equation}
    p\left(\ystate | \joint^H, \joint^{AI}, p\right) = p.
\end{equation}


\subsection{Decisions under miscalibrated probabilities}
Of course, in practice, model confidence scores will not necessarily equal the true probabilities. One reason is unavoidable sampling error from finite data. Calibration requires sample complexity exponential in the number of classes~\citep{zhao2021calibrating} in multi-class classification. When perfect calibration is impossible, a question arises: Can a decision maker do better than acting as if the predictions are true if they are given information about the calibration error bound? 


Calibration error measures how ``far'' a predictor is from perfect calibration. For decision purposes, the hope is that calibration error captures the loss a decision maker incurs if they trust miscalibrated predictions. Thus, when the decision maker is told predictions have a low calibration error, she can trust the prediction without worrying about incurring high loss on the decision task. 
Recent theoretical work explores the implications for decision making of popular calibration error metrics.
For example, Expected Calibration Error (ECE), which captures the weighted average of prediction errors, has been a popular choice of metric within machine learning research for optimization and benchmarking due to its mathematical simplicity (e.g., \citet{guo2017calibration, minderer2021revisiting}). ECE upper bounds the decision loss for every decision task with bounded payoff~\citet{kleinberg2023u}. 
However, ECE is only a loose upper bound of decision loss; \citet{hu2024predict} propose an alternative calibration error that directly relates to and tightly upper bounds decision loss. For the purposes of our discussion, the point is that given an error measure that sufficiently captures decision loss, it becomes straightforward to anticipate decision loss given expected miscalibration. 


\subsection{Takeaways}
The use of calibrated predictions in decision making is complicated by the decision maker having private signals or knowledge of calibration error. However, even under these constraints, it is relatively straightforward to predict how a rational decision maker will use the predictions as a result of the alignment between good decision making and posterior probabilities. As we discuss next, this is not the case when the goal instead is to provide calibrated prediction sets.



\section{Decisions from conformal prediction sets}
\label{sec:conformal}


We now consider how the decision maker uses a conformal prediction set. In contrast to the well-defined use of calibrated probabilities by a rational decision maker, a signal that takes the form of a conformal prediction set ($\signal = \cset(\xnew=\xb) \subset\mathcal{Y}$) provides no measure of which elements of the set are more or less credible. This presents a challenge to assessing their value to decision makers, even in the simpler case where the decision maker does not possess private signals.

Informally, we might distinguish a few broad strokes ways in which a decision maker responds to a prediction set. Perhaps they use it to \textit{verify} their own judgment about the best course of action. For example, some participants who used prediction sets in an AI-advised image labeling task studied by \citet{zhang2024} described coming up with their own judgment on the correct label, then checking whether it was included in the set. Another approach reported by participants from the same study used the set to \textit{constrain} the labels they considered as they assessed the instance (in this case, an image to be classified). Similarly, \citet{lu2022fair} characterized two strategies for using prediction sets based on interviews with clinicians: to rule out critical conditions that would make a treatment plan risky (such as giving a certain medication to a patient for which it might cause complications), and to rule in critical conditions to distinguish which patients should remain in the hospital.  

On the other hand, despite prediction sets not including a measure of credibility over elements, we can lower bound the loss attainable by any decision maker by assuming a Bayes rational agent who possesses information about the joint distribution over signals and states, and uses the joint distribution to update its beliefs upon viewing a set.

Below, we formalize a set of strategies that a decision maker might use to select a label from a prediction set, distinguishing between strategies that assume knowledge of the joint distribution over signals and states, strategies that combine the decision maker's prior knowledge with the set in a non-Bayesian way, and strategies where the set constrains the decision maker's action. 

\begin{table}[h]
    \centering
    \footnotesize
    \begin{tabularx}{\textwidth}{XXX}
         \textsf{Benchmark} & \textsf{Signals} & \textsf{Belief updating or decision strategy} \\ \toprule
            \cellcolor{lightgray}\textbf{Oracle} &\cellcolor{lightgray} & \cellcolor{lightgray}
\\
         \specialrule{0.25pt}{0.25pt}{0.25pt}
         
         Bayesian DM who knows \newline $\joint \in P(\dataspace \times \ystatespace)$ & $\signal = \{\xb, \mathcal{\cset}(\xb)\}$ & $p(\ystate|\signal) = \frac{\joint(\xb, \ystate)}{\sum_{\ystate'\in\ystatespace} \joint(\xb, \ystate')}$ \\     \specialrule{0.25pt}{0.25pt}{0.25pt}
        Bayesian DM who knows \newline $\joint \in P(2^\ystatespace \times \ystatespace)$ & $\signal = \mathcal{\cset}(\xb)$ & $p(\ystate|\signal) = \frac{\joint(\mathcal{\cset}(\xb), y)}{\sum_{\ystate'\in\ystatespace} \joint(\mathcal{\cset}(\xb), \ystate')}$\\
         \toprule
            \cellcolor{lightgray}\textbf{Use set to verify a priori judgments} & \cellcolor{lightgray} & \cellcolor{lightgray}\\
         \specialrule{0.25pt}{0.25pt}{0.25pt}
        Non-Bayesian DM with prior and misspecified beliefs \newline $\joint\in\distover{\ystatespace}$ & $\signal = \mathcal{\cset}(\xb)$ & 
        \(p(\ystate|\signal) = \) \newline
        \(\begin{cases}
          \frac{p(\ystate)}{\sum_{\ystate' \in \signal} p(\ystate')} \cdot (1-\alpha) & \ystate \in \signal \\
          \frac{p(\ystate)}{\sum_{\ystate' \in \ystatespace \setminus \signal} p(\ystate')} \cdot \alpha & \ystate \not\in \signal 
        \end{cases}\)
         \\ \specialrule{0.25pt}{0.25pt}{0.25pt}
        Associative non-Bayesian DM with \newline prior $\joint(\ystatespace)$ and similarity function \newline $d: \ystatespace \times \ystatespace \rightarrow \mathbb{R}$ & $v = \{y: y\in \ystatespace \wedge \exists y' \in \mathcal{\cset}(\xb)$ \newline $\text{ such that} \ d(y,y') \leq b\}$  & Same as non-Bayesian DM  \\ 
        \toprule
        \cellcolor{lightgray}\textbf{Use set to constrain decision options} & \cellcolor{lightgray} & \cellcolor{lightgray}\\
         \specialrule{0.25pt}{0.25pt}{0.25pt}
        Uncertainty suppressing DM divides probability over set & $\signal = \mathcal{\cset}(\xb)$ & $p(\ystate|\signal) = \begin{cases}
             \frac{1}{|\signal|} & \ystate \in \signal \\
             0 & \ystate \notin \signal
        \end{cases}$ \\ \specialrule{0.25pt}{0.25pt}{0.25pt}
        Conservative risk-averse DM & $\signal = \mathcal{\cset}(\xb)$ & $\action^*(\cset) = $ \newline $\argmin_{\action'\in \actionspace}\max_{\ystate\in \cset} \score(\action', \ystate)$
        \\ \toprule
\cellcolor{lightgray}\textbf{Baseline with no signals} &\cellcolor{lightgray} & \cellcolor{lightgray} \\
        \specialrule{0.25pt}{0.25pt}{0.25pt}
        Bayesian DM who knows only prior \newline $\joint\in\distover{\ystatespace}$ & No signal & $p(\ystate|\signal) = \joint(\ystate)$ \\ \toprule
    \end{tabularx}
    \caption{Overview of signals and associated strategies for using prediction set information. The ranking of benchmarks denoting the best possible performance in each scenario (Equation~\ref{eq:benchmark}) is partially predictable. Row 1 dominates row 2 and row 2 dominates all other rows. However, where the non-Bayesian decision makers fall relative to the Bayesian decision maker with only the prior (bottom row) depends on the structure of the specific setting.}
    \label{tab:my_label}
\end{table}

\subsection{Decision maker with oracle access to joint distribution over signals and states}
We first consider, What is the best possible expected performance of any decision maker who is provided with prediction sets as decision support? We define two benchmarks that assume oracle access to joint probability information connecting the signals and the state. The first assumes the decision maker has access to joint information linking the state to the features. The second assumes that the decision maker has access to joint information linking the state to the prediction sets.

\subsubsection*{Lower bound: Bayesian decision maker who knows joint distribution over features and outcomes}

A natural signal structure to consider presents the decision maker with instance features $\xb$ and the corresponding prediction set $\signal = \{\xb, \mathcal{\cset}(\xb)\}$ from some conformal algorithm. 

When the instance information $\xb$ is included in the signal, then the best attainable performance by any decision maker with signals of the form $\signal = \{\xb, \mathcal{\cset}(\xb)\}$ is that of the Bayes rational decision maker who knows $\joint \in \distover{\dataspace \times \ystatespace}$. However, once joint information about the instance features and state are known, prediction sets can offer no additional information. For a decision maker with knowledge of the joint distribution over $\dataRV$ and $\ystateRV$, the probability $\joint(\xb,\ystate)$ already contains all relevant information needed to arrive at the Bayes posterior beliefs~\ref{eq:posterior}, thus the benchmark is a strict lower bound that is unaffected by the removal of $\mathcal{\cset}(\xb)$ from the signal. This benchmark helps illustrate that for prediction sets to make sense as a form of decision aid, we must not believe that the decision maker is well informed of the relationship between $\dataRV$ and $\ystateRV$. In other words, whenever we believe that prediction sets will be informative to the decision maker, we should not expect this benchmark to be attainable.

\subsubsection*{Bayesian decision maker who knows joint distribution over sets and outcomes}

If we think decision makers will rely primarily on the prediction sets rather than their knowledge of the relationship between $\dataRV$ and $\ystateRV$, the best attainable performance is captured by a lower bound representing the maximum attainable score of a Bayes rational decision maker with access to the joint distribution over prediction sets and outcomes, but \textit{not} the instance features. Here, the signal space $V = 2^\ystatespace$ and $\joint$ assigns a probability $\joint(\signal, \ystate)$ to each label $\ystate$ and each possible prediction set $v$ (a conjunction of indicators of whether each label is in the set).

For example, imagine there are three states: $\statespace = \{\mstate_A, \mstate_B, \mstate_C\}$. Imagine that $\joint$ assigns prior beliefs $\Pr(\mstate_A) = \frac{1}{2}, \Pr(\mstate_B) = \frac{1}{3}, \Pr(\mstate_C) = \frac{1}{6}$.
Say the decision maker observes a particular set $v = \mathcal{\cset}(\xb) = \{\mstate_B,\mstate_C\}$. They update their beliefs using Bayes rule to arrive at posterior beliefs $\joint(\mstateRV|\signal)$ (Equation~\ref{eq:posterior}), and choose the utility-maximizing action (Equation~\ref{eq:opt}).

One challenge with this benchmark is that it is vulnerable to overfitting as the label space grows (as is the prior benchmark when the feature and label spaces are large).  
Additionally, while this lower bound may be more attainable to human decision makers than the previous one, it also challenges intuitions we may have about prediction set use in a few ways.
First, the assumption that the 
decision maker has no knowledge of how the features $\dataRV$ associate with the state $\ystateRV$ can seem unrealistic in settings where experts are assisted by models. Here, the contradiction is that it makes little sense to have humans in charge of final decisions if we don't think they have any direct knowledge of how instance features correlate with outcomes. 

Second, the coverage guarantee that is the focal point of the conformal literature is only indirectly relevant to this decision maker.
The benchmark is akin to imagining a decision maker whose knowledge consists of having seen the prediction sets and ground truth labels for instances in the calibration set. Coverage information is implicit in the joint distribution they have access to, but we make expect human decision makers to rely more directly on the coverage guarantee in how they use the set.


\subsection{Decision maker uses set to ``verify'' a priori judgments}
If we believe a decision maker uses the prediction set to confirm that a label they had in mind is correct, then Bayesian updating is violated. Believing that a prediction set meaningfully signals which states are possible also implies that the decision maker has false expectations about how sets are generated. However, such strategies may better capture intuitions about how human decision makers will use prediction sets than assuming oracle access to joint probabilities.

\subsubsection*{Non-Bayesian decision maker with prior and misspecified beliefs about sets}
\label{sec:misspecified}

There is precedent in information economics and game theory for analyzing belief updating from sets of possible states. Signals consisting of partitions of the statespace appear in well-known constructions like the agreement theorem of \citep{aumann2016agreeing}.
Under this interpretation, a prediction set denotes which states are possible.

Assume a decision maker who has knowledge of the prior distribution over states $\joint(\ystateRV)$.
Upon viewing a signal of the form $\signal = \mathcal{\cset}(\xb)$, 
for each possible label $y \in \ystatespace$, the decision maker updates their beliefs by noting whether the label is in the set, normalizing its prior probability using the relevant subset, then multiplying it by the appropriate probability: 

\begin{equation}
p(\ystate|\signal) = 
\begin{cases}
  \frac{\joint(\ystate)}{\sum_{\ystate' \in \signal} \joint(\ystate')} * (1-\alpha) & \ystate \in \signal \\
  \frac{\joint(\ystate)}{\sum_{\ystate' \in \ystatespace \setminus \signal} \joint(\ystate')} * \alpha & \ystate \not\in \signal 
\end{cases}
\label{eq:setastrue}
\end{equation}

\noindent where all labels have strictly positive prior probability and $\ystatespace \setminus \signal$ denotes states not in the prediction set. 

For example, given the decision problem above with prior $\Pr(\mstate_A) = \frac{1}{2}, \Pr(\mstate_B) = \frac{1}{3}, \Pr[\mstate_C] = \frac{1}{6}$ and signal $v = \{\mstate_B,\mstate_C\}$, the decision maker selects the utility optimal action given posterior beliefs $\Pr(\mstate_B|v) = \frac{2}{3}\times0.95$, $\Pr(\mstate_B|v) = \frac{1}{3}\times0.95$, and 
$\Pr(\mstate_A|v) = 0.05$.

When the decision maker lacks information about the generating process that produces a prediction set, the setting becomes a decision problem under misspecified beliefs, as studied in economics, e.g.\ \citet{fudenberg2017active, heidhues2018unrealistic}. Applying this strategy to conformal prediction sets implies that the decision maker's beliefs are misspecified in the sense that they believe the data-generating process generates prediction sets from partitions, i.e., non-overlapping subsets. In practice, it is difficult to motivate developing conformal algorithms with this criterion, as it further restricts an algorithm's ability to achieve desirable properties like adaptiveness through varying set size. We include this strategy here however because it proposes a plausible form of belief misspecification, resulting in a decision rule that depends only on the set and the prior distribution of the states.



\subsubsection*{Associative non-Bayesian decision maker with prior and misspecified beliefs about sets}
\label{sec:associative}
In many domains with large label spaces to which conformal prediction has been applied, the labels are naturally organized hierarchically. 
For example, the decision utility of a prediction set for medical decision tasks can be modeled to account for the number of categories of diseases the set captures as defined by a medical hierarchy~\citep{cortes2024decision}. 
Image classification is scaffolded by a hierarchy of concepts taken from WordNet's semantic ontology~\citep{deng2009imagenet}, where the shortest path in the hierarchy is sometimes used as a proxy for image similarity. 
Recent empirical study of the use of prediction sets for AI-advised image labeling suggested that one way people may make decisions from a set uses semantic similarity to transfer probability to labels that are close to those included in the set, but not included themselves~\citep{zhang2024}.

For example, assume a distance function $d$ that assigns a score $d: \ystatespace \times \ystatespace \rightarrow \mathbb{R}$ to each pair of labels, minimized at $d(y,y) = 0$. Given a distance $b$ beyond which two labels are no longer considered closely related, we replace $v$ with the set of all labels that are sufficiently similar to at least one label in $v: v' = \{y: y\in \ystatespace \wedge \exists y' \in v \ s.t. \ d(y,y') \leq b\}$. Equation~\ref{eq:setastrue} above may then be applied by an associative decision maker by replacing $v$ with $v'$. 

\subsection{Decision maker uses set to constrain decision}
A final set of strategies assumes that the prediction set guides which labels a decision maker considers. This category is most closely related to the intuition that prediction sets reduce cognitive processing on the part of the decision maker. Both strategies discussed below are also amenable to associative variations where proximity to labels in the set results in additional labels being considered.

\subsubsection*{Uncertainty-suppressing decision maker who divides probability equally over states in set}
\label{sec:uncertaintysuppress}

Prior empirical work on the use of prediction sets in image labeling found that humans achieve higher accuracy when restricted to choosing a label from within prediction sets~\citep{straitouri2023designing}. If we expect people to suppress uncertainty in decision making, e.g., by rounding large probabilities up rather than maintaining the small possibility of alternatives, one plausible variation on the strategy discussed above is to assume that the decision maker will simply randomly choose an element from those in the set given beliefs: 

\begin{equation}
\dist(\ystate|\signal) = \begin{cases}
     \frac{1}{|\signal|} & \ystate \in \signal \\
     0 & \ystate \notin \signal
\end{cases}
\label{eq:divideprob}
\end{equation}

\subsubsection*{Cost-constrained Bayesian decision maker who uses prediction set as prior}
\label{sec:costconstrained}
A prediction set may reduce cognitive processing for the decision maker in situations where attention is scarce and prohibits considering all possible labels. This might be the case, for example, if further information can be gathered to support a decision (e.g., about a patient who needs to be diagnosed) but the space of possible information to collect is large, for example because there are a large number of symptoms to check for or tests that could be requested. Alternatively, a decision maker might rely on a prediction set to provide a prior on what labels to keep in mind as they process a high dimensional representation of an instance like an image of a patient's condition. For example, some study participants who used prediction sets for the AI-advised image labeling in \citep{zhang2024} reported viewing the set for possibilities before looking carefully at the image.

A rational inattention model from information economics provides a theoretic foundation for well-studied behavioral patterns \citep{matvejka2015rational}. This model assumes a decision-maker facing a 
discrete choice situation in which they do not
know the values of the available options, but have an opportunity acquire and process information at some cost ~\citep{stigler1961economics, morgan1985optimal}. The model quantifies the cognitive cost of information for a rational agent by the entropy reduction in their beliefs. 
In a rational inattention framework \citep{sims2003implications}, instead of solving the payoff-maximizing decision problem in Equation \ref{eq:opt}, the agent optimizes for the expected payoff after subtracting the cognitive cost. 

Consider the uniform prior as an example. Under the rational inattention framework, in tasks with $|\ystatespace|$ labels (e.g. $\sim 1000$ labels in \citep{zhang2024}), the agent suffers a cognitive cost of $O(|\ystatespace|)$ to figure out the true answer without the help of a conformal set, where the cost is not necessarily recovered by the payoff gain from figuring out the true label. This results in \textit{rational} selection of a less informative but less costly signal. 
For example, an agent might check a few but not all labels. Checking a few labels is less informative but also less costly than figuring out an answer from all $\sim 1000$ labels. 
A conformal set reduces this cognitive cost by effectively refining the posterior over the set for free, allowing the agent to allocate cost to obtaining information over states in the set. 

\subsubsection*{Conservative decision maker who maximizes worst-case utility}
\label{sec:maxmin_roth}
Thus far we have primarily considered risk neutral decision-makers whose objective is to choose the action that minimizes their expected loss: $\argmin_{\action\in \actionspace}\expect_{\mstateRV\sim \joint(\cdot|\signal)}[\score(\action, \mstateRV)]$. 
However, unlike point predictions, conformal sets do not provide agents with quantitative risk assessments, suggesting a connection to maxmin (risk-averse) utility optimization. Recent work by \citet{kiyani2025decision} makes progress on the question: How should a decision maker who has access to only a prediction set choose actions to maximize risk-averse utility? They show that, considering all data-generating processes $\joint(\dataRV, \ystateRV)$ that are compatible with the set coverage $\alpha$, if the optimization goal of the decision-maker is to maximize worst-case expected utility, the optimal strategy amounts to a simple max-min strategy over the predicted labels in the set.

Formally, without any knowledge about the data-generating process, the decision-maker is constrained to select an action $\action: \cset\to \actionspace$ as a function of the conformal set. Fixing a data-generating process $\joint(\dataRV, \ystateRV)$, an action policy $\action(\cdot)$ is evaluated by the expected $\alpha$ quantile of loss $l(\dataRV)$:
\begin{align}
\score^*(\action,\joint) =
\min_{\nu(\cdot)} & \qquad \expect_{(\dataRV, \ystateRV)\sim \joint} [
\quantile_{\alpha}[\score(\action(\cset(\dataRV)),\ystateRV)]] 
\end{align} 
This $\alpha$ quantile of loss serves as a robust upperbound on the loss for any sample instance with high probability $1-\alpha$. 


A risk-averse decision-maker optimizes $\nu^*(\action, \joint)$ for any data-generating process $\joint$ with compatible coverage guarantee. We write the set of compatible $\joint$ as $\compatiblejoint$, i.e.\ for any $\joint\in \compatiblejoint$, 
\begin{equation*}
    \Pr_{(\dataRV, \ystateRV)\sim \joint}[\ystateRV\in \cset(\dataRV)]\geq 1-\alpha.
\end{equation*}
The decision-maker solves the risk-averse optimization
\begin{equation}
\label{eq: conformal risk averse opt}
    \min_{\action}\max_{\joint\in \compatiblejoint}\score^*(\action, \joint)
\end{equation}
\citet{kiyani2025decision} shows that, for $\alpha\leq \frac{1}{2}$, the optimal solution to \Cref{eq: conformal risk averse opt} is the simple min-max strategy over labels in the set, i.e.\ 
\begin{equation*}
    \label{eq:kiyani_set_optimal}
    \action^*(\cset) = \argmin_{\action'\in \actionspace}\max_{\ystate\in \cset} \score(\action', \ystate).    
\end{equation*}

This finding specifies optimal usage of prediction sets in a risk-averse decision scenario. The decision maker can simply treat the labels in the prediction set as defining possible realizations of the state, and apply what they know about the loss function to select the label in the set with the best worst-case loss.

\subsection{Takeaways}
These examples help illustrate two important takeaways. 
First, the fact that we cannot specify a single best procedure for a risk-neutral agent who uses prediction sets highlights the underspecification of ``good'' decision making with conformal prediction, including in the simplified case where we do not expect the decision maker to have access to additional information over the model. 
This is not necessarily a reason to avoid using prediction sets, as there is no guarantee that human decision makers will engage in idealized use of an uncertainty quantification when it is well defined. However, that this aspect of conformal prediction use has largely gone unacknowledged in the literature suggests that there is work left to do to characterize successful integration of conformal prediction in human-facing decision workflows.

On the other hand, assuming that decision-makers will be risk-averse may be reasonable in some settings where prediction sets attract interest, such as medical diagnosis or treatment decisions. Even if real human decision-makers cannot confidently rank actions by their worst-case loss upon first encountering a set, in some cases subsequent information gathering may be enough to reduce ambiguity, making \ref{eq: conformal risk averse opt} a reasonable strategy to assume.

Second, the above examples illustrate how our ability to predict their usefulness depends on what we think human decision makers know and what sort of constraints we believe they are under. To design prediction sets that are useful for real human decision makers, these aspects of a decision scenario must be considered explicitly.

\section{The meaning of quantified uncertainty is its use}
\label{sec:discussion}

We have demonstrated some ways in which conformal prediction sets are elusive as a form of decision support. How should ambiguity about how prediction sets are best used by decision makers influence how we see the value of conformal prediction for quantifying uncertainty? What considerations should guide how conformal prediction is implemented in practical decision workflows, and what gaps are most important to address through future research?

One perspective on these tensions is that we should expect the meaning of quantified uncertainty to lie in its use. This perspective has several implications. 
Rather than prioritizing the ability to prove theoretical guarantees and looking to domain-specific decision processes primarily for datasets or use cases, we should be considering how different forms of guarantees align with the needs and sensitivities of human decision makers throughout method development. We should study the use of conformal inference from the perspective of developers and decision makers, and expect its value in practice to sometimes lie beyond our attempts at formal characterization. 
We reflect below on what this suggests for research aimed at improving the utility of prediction sets for human decision makers, and more broadly how it affects how we ``price in'' conformal prediction relative to alternative techniques.

\subsection{A human-centered research agenda for conformal prediction}

\vspace{3mm}
\noindent \textbf{Understanding and guiding human decision makers' use of prediction sets}. 
Our work suggests that despite its popularity in practice and theoretical research, we lack a deep understanding of conformal prediction as a human-facing form of decision support. 

One way to approach empirical study of prediction set use is to compare observed human performance with prediction sets to benchmarks based on the strategies we describe (Section~\ref{sec:conformal}). We associate this approach with the general practice of statistical model checking~\citep{hullman2021designing}, where in this case the goal is to identify which of several models appears to best explain human performance and to observe deviations that remain to be explained. There is precedent for designing experiments to discern possible mechanisms behind human decision performance under uncertainty in the data visualization literature (e.g.,~\citep{kale2020visual,wu2023rational}).
A focus on testing user performance against expectations given different strategies provides a more robust basis for further theory development than approaching uncertainty quantification for humans as if they are a blackbox to optimize (e.g., by running a grid search over different hyperparameters like coverage level to find the values that lead to the best performance). 
We should expect human performance and use to vary as we vary aspects of the decision problem, loss function, and problem setting. The question is how changes in a decision problem impact reliance on different strategies for using sets. Hence, we should avoid drawing general conclusions about the level of coverage or particular algorithm that will work best for human decision makers more broadly from any single experiment. 


\vspace{3mm}
\noindent \textbf{Incorporating expert knowledge into predictive uncertainty design}.
Given that human experts are often paired with AI models because they are thought to have complementary information, it is critical for future research to acknowledge information complementarities and pursue methods for characterizing and designing for the particular information conditions under which predictive uncertainty will be presented~\citep{corvelo2024human,guo2024unexploited}. This means recognizing the potential disparity between probability guarantees defined on information available to a model and those that are directly relevant to a human with additional information. Whenever we believe the human decision maker may possess information beyond that encoded in the model, we should expect probabilities based only on the model's available information to be  wrong from the decision maker's perspective.

Related to this, organizations deploying AI models to assist human decision makers often choose to keep the human in the loop because of 
potential flexibility in their decision strategies in the face of shifts in the decision-relevant distribution. Understanding the robustness of coverage achieved by standard conformal algorithms to common forms of shift, and how well humans can pick up on shifts even if a model may not, is a well-motivated direction for future work.

\vspace{3mm}
\noindent \textbf{Reconsidering conformal guarantees from a human perspective}. 
The conformal prediction literature prioritizes coverage guarantees over other properties that may be desirable to human decision makers, such as low decision loss or properties of set size. \citet{kiyani2025decision}, for example, maximize utility subject to coverage guarantees. In practice, however, coverage may be less important to decision makers, suggesting an inversion of this approach where coverage is maximized subject to minimum utility or other requirements that are important to human decision makers, like the information captured by set size. For example, work by \citet{straitouri2023improving} asks what coverage level leads to the most effective human decisions. 
Empirical work aimed at re-prioritizing properties of prediction sets for human alignment can be pursued in tandem model checking approach to experimentation described above, to allow for solutions to be adapted to slightly different scenarios. 
This is because empirical estimates of the parameters maximize expert performance learned without also attending to underlying mechanisms may be brittle to slight changes in the decision problem or expert population.

\vspace{3mm}
\noindent \textbf{Aligning uncertainty quantification theory with human preferences}
Related to our point about re-prioritizing properties of prediction sets, future research should attempt to better understand and integrate aspects of decision makers' preferences for uncertainty quantification in conformal theory. These include set size, which is usually achieved through heuristics, and the ``shape'' that sets can take, particularly in higher dimensions where designers must constrain the structure (e.g., only considering contiguous regions or ellipsoids in a high dimensional space).

Part of these efforts will include identifying effective ways to elicit human preferences for uncertainty quantification. While some recent work demonstrates how to incorporate a domain-specific, downstream decision loss function into the generation of prediction sets~\citep{cortes2024decision}, this information is assumed to be given as input. 
To align predictive uncertainty with human needs and values in arbitrary domains, we will also need elicitation methods that can be brought to bear in arbitrary domains where decision problems and associated preferences have not yet been fully characterized. 
Another direction is to pursue interactive methods that help experts steer the prediction sets that are generated to meet their needs, such as by enabling them to ``query'' a set for potential outcomes that are most relevant to their information needs.


\subsection{Pricing in conformal prediction}
Statistical theory is sometimes said to be the theory of applied statistics. Conformal inference demonstrates how established theory in rank-based statistics can be applied to modern problems of quantifying prediction uncertainty. The attractiveness of prediction sets to human decision makers warrants future research aimed at increasing the alignment between conformal methods and experts' needs and preferences. Yet, we should expect it to remain difficult to fully characterize the value of conformal prediction relative to alternative approaches to uncertainty quantification. We discuss several considerations that may be important in practice but remain difficult to formally ``price in,'' and advocate for acknowledging these considerations in practice despite their slipperiness as theoretical constructs.

\subsubsection{The twoishness of post hoc calibration}

Offline calibration methods can often be described as relying on two models: the original black-box predictive model and the model of the frequency distribution of the predictions learned from calibration data.
Split conformal prediction combines an arbitrary predictive model whose output we use to derive a non-conformity score as our data model (or likelihood) with a second model defined by the sampling distribution of the non-conformity scores. This in not unique to conformal prediction: many statistical methods benefit from a certain ``twoishness'' in their construction~\citep{gelman2011}. For example, a common pattern occurs in classical statistical analysis, where one method is used to produce estimates (e.g., a multivariate regression to estimate effects and compute standard errors) and another is used to check if the first method makes sense (e.g., predictive checks or significance tests on the coefficients).
When it comes to conformal prediction, the benefits of combining these two models are obvious: we are absolved of having to understand the relationship between the model's training distribution and the deployment (test) distribution, and can ignore that the original model may be misspecified to instead focus on deriving prediction probabilities that hold on calibration data we choose.

At the same time, twoishness tends to coincide with limitations inherent to the modeling task. For conformal prediction, the ability to rely on new data to achieve coverage is most useful when we find ourselves in the limited position where we cannot retrain the predictive model despite a mismatch between the training and deployment context. 
A risk in such situations is that the gaps that remain when we combine multiple models are de-prioritized relative to the "solution" that the combination provides, such that practitioners fail to acknowledge critical contingencies.

As an analogy, consider the standard bootstrap, a distribution-free technique for quantifying uncertainty that resembles conformal prediction in substituting simple computations for explicit distributional assumptions.
In standard non-parametric bootstrapping, we have some observed data $\mathcal{D} = (X_{1}, \dots, X_{n}) \sim P$ and wish to quantify the uncertainty in our plug-in estimate of a ``parameter'' of $P$, $\hat{\theta}_{n}$.
Constructing valid $1-\alpha$ confidence intervals on our sample estimate means approximating the exact finite sample distribution of a test statistic involving the estimator, for example:

\begin{equation}
F_{n}(t) = \mbox{Pr}(\sqrt{n}(\hat{\theta^{*}} - \hat{\theta}_{n}) \le t)
\end{equation}

This is accomplished by estimating $F_{n}$ using an empirical approximation $\hat{F_{n}}$ conditional on $D$, which is treated as fixed. Specifically, the Monte Carlo approximation $\bar{F}$ is used:

\begin{equation}
   \bar{F}(t) =  \frac{1}{B} \sum_{j=1}^{B} I(\sqrt{n}(\hat{\theta^{*}_{j}} - \hat{\theta}_{j}) \le t)
   \label{eq:bootstrap_iterations}
\end{equation}
Key ideas behind the bootstrap include the observation that an unknown probability distribution can be replaced with an empirical estimate (the substitution principle), and the use of simulation to avoid analytically calculating properties of an estimator~\citep{davison2003recent}. 
Related to twoishness, when we acknowledge the distinction between the model under which we define the estimator $\hat{\theta_{n}}$ and the model of the sampling process, it becomes clearer that while the bootstrap allows us to minimize the error or uncertainty from the Monte Carlo approximation (by setting a high enough number of iterations $B$ in \ref{eq:bootstrap_iterations}, 
we cannot reduce our uncertainty about how well our particular sample $\mathcal{D}$ represents $P$, and therefore our uncertainty about how well $\hat{F_{n}}$ approximates $F_{n}$. It would be a mistake to adopt the bootstrap in any particular setting without considering how the irreducible uncertainty due to having only a single sample might affect our goals, through, for example, the influence of outliers on the target statistic. 
 
Similarly, in research on conformal prediction, we should not let our ability to prove guarantees about conformal methods overshadow acknowledgement that whenever we bring in calibration data after model training, any ``guarantees'' are contingent on the fitness of the calibration data as a representation of the ground truth generating process and on our ability to imagine taking hypothetical random samples of calibration and test data.
At minimum, this suggests emphasizing the conditionality of guarantees when describing methods like conformal prediction to practitioners and other non-theorists who may not fully grasp the underlying assumptions (of a static generating process, the relevance of asymptotic theory, and so on).

\subsubsection{The allure of thoughtless statistics}
On the other hand, we should also expect approaches to quantifying uncertainty to sometimes have value that falls outside the bounds of available theory.  
For example, the existence of second-order asymptotic guarantees on coverage does not necessarily help us understand the popularity of the standard bootstrap. 
Instead, the widespread appeal of the bootstrap in practice likely owes much to its generality of construction.
By avoiding distributional assumptions and substituting computation for analytical calculation, the bootstrap promises quantified uncertainty with assumptions that are implicit---but its popularity surely also derives from its twoishness, which (a) incorporates additional information about the sampling procedure that is not included in the estimator being bootstrapped, and (b) gives additional flexibility to the user. Bootstrap theory cannot comment on its perceived value as an exploratory tool, even in settings where use of the bootstrap is not formally justified (e.g., where observations are correlated to some extent, or outputs consist of graphs or trees; \citep{holmes2003bootstrapping}). Because the procedure is simple and seemingly well understood, applying the bootstrap can still provide some intuition into the variance in a procedure when the typical conditions do not hold.

Similarly, split conformal prediction may appeal over more robust uncertainty quantification approaches that ensemble several models or refine the base predictive model because of its procedural simplicity.
Even when we don't expect conformal coverage guarantees to hold exactly as stated (e.g., because of data shifts or constraints like spatial or temporal correlations, or additional knowledge a human possesses), relative information about model uncertainty captured in set size   (e.g.,~\citep{angelopoulos2021uncertainty}) may still be valuable in many settings.

We should expect decision makers to perceive their options differently when it comes to approaching decision making with access to prediction sets versus labels with predicted probabilities. For example, we would expect uncertainty suppressing decision makers who round predicted probabilities (Section~\ref{sec:uncertaintysuppress}) to diversify their actions more with prediction sets than predicted probabilities. There is a sense in which prediction sets provide a more concrete depiction of possibilities, such that findings that behavioral decision makers benefit from discrete framings of probability may apply~\citep{hullman2015hypothetical,kay2016ish,kale2020visual}. 
It is also possible that seeing more labels, as opposed to seeing only the model's top prediction or top couple predictions, has value from the standpoint of encouraging broader thinking about possible outcomes, or from a debugging perspective, where a decision maker may be more likely to consider sources of epistemic uncertainty upon seeing a prediction set compared to a top model prediction. These possibilities are not easily ``priced in'' and consequently get overlooked in theoretical discussions.

At the same time, it is important not to assume that seemingly large differences in the philosophy behind different expressions of predictive uncertainty means that the decisions they induce will be equally distinct. Effect sizes in existing empirical work comparing human decisions under prediction sets versus alternatives like top-$k$ sets based on heuristics like softmax scores are relatively small, even as researchers study settings where we might expect conformal sets to have advantages~\citep{cresswell2024,cresswell2024conformal,zhang2024}. \citet{zhang2012ubiquitous}, for example, find that even in on out-of-distribution images where prediction sets showed the biggest advantage of top-$k$ presentations with softmax values, the difference in decision accuracy of the human-AI team was only 4 percentage points. Small effects (e.g., on the order of a few percentage points difference in probability perception) are also common in research on uncertainty visualization~\citep{kale2020visual}. 
Reflecting on how to rigorously quantify uncertainty can play a critical role in shaping future data collection, evaluation design, and other components of a modeling-to-decision pipeline, but once put into practice, the benefits of a chosen quantification can be difficult to demonstrate.

\subsubsection{Second-order uncertainty: What can't be learned can't be priced}
Discussions of sources of uncertainty often distinguish \textit{aleatoric uncertainty}---uncertainty in an outcome due to inherent randomness in a process---from \textit{epistemic uncertainty} due to a lack of knowledge that could in theory be reduced. While calibration can assure good decisions under aleatoric uncertainty under certain assumptions, its resolution of epistemic uncertainty is arguably less satisfying. However, epistemic uncertainty cannot be ``priced in'' to the loss minimization framework in which many predictive uncertainty techniques are evaluated.

Returning to multiclass classification as an example, a classifier $\model$ trained to minimize loss using a set of labeled training data can be thought of as producing predictions at different levels, which are evaluated using different empirical values. We can evaluate the ``level 0'' prediction $\ypredict = \model(\dataRV=\xb)$ of the label against the true label using 0/1 loss. We can evaluate the ``level 1'' prediction of the conditional distribution $\modelprob(\cdot|\dataRV=\xb)$ against the empirical distribution of labels. 

Aleatoric uncertainty captures the fact that even if we were given full knowledge of $p(\cdot|\dataRV=\xb)$, we cannot predict with certainty due to inherent randomness in the generating process. 
The ideal case for deploying a statistical predictor to make decisions is when we wish to minimize decision loss on average, and have access to a large amount of data that we are confident reflects the target generating process. In such cases, we are primarily concerned with aleatoric uncertainty. 

In contrast, epistemic uncertainty deals with the possibility that the learned probabilistic predictor $\modelprob$ associated with classifier $\model$ differs from the ground truth $p$. Distribution-free post hoc calibration of prediction uncertainty promises to ``fix'' the gap between $\modelprob$ and $p$, but offers the decision maker no information to estimate epistemic uncertainty, i.e., how different the prediction $\modelprob(\cdot|\dataRV=\xb)$ may be from the ground truth. 
However, 
because we can't observe samples from the second order distribution $\mathbb{P}(\mathbb{P}(\ystate|\dataRV=\xb))$, we cannot learn a faithful predictor of higher order epistemic uncertainty within a loss minimization framework~\citep{bengs2022pitfalls}. This is formalized by \citet{bengs2023second}, who relate it to the impossibility of defining a proper scoring rule. Learning to predict conditional distributions $\modelprob(\cdot|\dataRV=\xb)$ using empirical risk minimization--i.e., identifying the empirical risk minimizer that minimizes loss on the training data---under a proper scoring rule means that the optimal predictor (the true risk minimizer or Bayes predictor) will coincide with the true conditional class distribution $\joint(\cdot|\dataRV=\xb)$. 
However, there is no loss function for comparing second-order predictions of $p(p(\cdot|\dataRV=\xb))$
with observations $\ystate$ such that minimizing that function on the training data ensures a second-order predictor will be ``honest'' in the sense of returning the prediction corresponding to its belief.

When we specify priors on model parameters in a Bayesian model, we explicitly express this uncertainty in our predictions. The
posterior predictive distribution $p(y_{rep}|y)$ marginalizes the distribution of $y_{rep}$, which we can think of as a newly drawn version of $y$ given $\theta$, over the posterior distribution of $\theta$ given $y$, which, by virtue of conditioning the prior, captures epistemic uncertainty stemming from our lack of knowledge of the model parameters. 
In a machine learning context, the posterior predictive distribution of a Bayesian neural network with latent variables captures uncertainty in network weights as well as in random noise and latent variables in data generation~\citep{depeweg2018decomposition}. 
However, Bayesian probability is not distinguishable in the Frequentist framework assumed by post hoc calibration methodologies.

Rather than arguing for one statistical paradigm over another, our point here is simply that there is always a risk to concluding that because some properties of uncertainty quantification cannot be incorporated into some particular framework, 
they cannot have value for decision makers in practice. 
Consider a situation where abstaining from making a decision in order to gather more information is possible. Offering the decision maker a probability distribution over the rate at which an event is expected to occur (as compared to a maximum likelihood point estimate) provides them a sense of how much risk is incurred by acting as if the maximum likelihood prediction is true versus holding out until more information is available. Similar to how we should consider what information is available to human decision-makers versus artificial agents when we design uncertainty quantification, we should also consider whether decision-makers would benefit from epistemic uncertainty that expresses the limitations of knowledge for a particular decision. This may be particularly desirable in situations where decision-makers are liable for individual decisions, such as about specific patients in a medical setting.

In conclusion, we should not forget that the meaning of quantified uncertainty as decision support for human experts often lies in its use. 
We should see the twoishness of a statistical method as a methodological ``seam'' reminding us that what a method provides is always a function of how its distinct component models combine. We should keep our eye on what lies beyond our ability to theorize, pursuing methods for better integrating human domain expertise and preferences and acknowledging when decision makers may desire forms of uncertainty that lie outside our chosen framework.  
Ultimately, we should not expect theory to capture everything valuable about a method, but we should continue trying to grow the reach of theory to include the practical questions. In other words, we should aim for theoretical statistics to be the theory of applied statistics without forgetting that there will always be gaps.






\bibliographystyle{plainnat} 

\bibliography{conformal}

\begin{thebibliography}{79}
\providecommand{\natexlab}[1]{#1}
\providecommand{\url}[1]{\texttt{#1}}
\expandafter\ifx\csname urlstyle\endcsname\relax
  \providecommand{\doi}[1]{doi: #1}\else
  \providecommand{\doi}{doi: \begingroup \urlstyle{rm}\Url}\fi

\bibitem[{\AE}gisd{\'o}ttir et~al.(2006){\AE}gisd{\'o}ttir, White, Spengler, Maugherman, Anderson, Cook, Nichols, Lampropoulos, Walker, Cohen, et~al.]{aegisdottir2006meta}
Stefan{\'\i}a {\AE}gisd{\'o}ttir, Michael~J. White, Paul~M. Spengler, Alan~S. Maugherman, Linda~A. Anderson, Robert~S. Cook, Cassandra~N. Nichols, Georgios~K. Lampropoulos, Blain~S. Walker, Genna Cohen, et~al.
\newblock The meta-analysis of clinical judgment project: Fifty-six years of accumulated research on clinical versus statistical prediction.
\newblock \emph{Counseling Psychologist}, 34\penalty0 (3):\penalty0 341--382, 2006.

\bibitem[Alur et~al.(2024)Alur, Laine, Li, Raghavan, Shah, and Shung]{alur2024auditing}
Rohan Alur, Loren Laine, Darrick Li, Manish Raghavan, Devavrat Shah, and Dennis Shung.
\newblock Auditing for human expertise.
\newblock \emph{Advances in Neural Information Processing Systems}, 36, 2024.

\bibitem[Angelopoulos and Bates(2023)]{angelopoulos2023conformal}
Anastasios~N Angelopoulos and Stephen Bates.
\newblock Conformal prediction: A gentle introduction.
\newblock \emph{Foundations and Trends in Machine Learning}, 16\penalty0 (4):\penalty0 494--591, 2023.

\bibitem[Angelopoulos et~al.(2021)Angelopoulos, Bates, Jordan, and Malik]{angelopoulos2021uncertainty}
Anastasios~Nikolas Angelopoulos, Stephen Bates, Michael Jordan, and Jitendra Malik.
\newblock Uncertainty sets for image classifiers using conformal prediction.
\newblock In \emph{International Conference on Learning Representations}, 2021.
\newblock URL \url{https://openreview.net/forum?id=eNdiU_DbM9}.

\bibitem[Angelopoulos et~al.(2024)Angelopoulos, Barber, and Bates]{angelopoulos2024online}
Anastasios~Nikolas Angelopoulos, Rina Barber, and Stephen Bates.
\newblock Online conformal prediction with decaying step sizes.
\newblock In \emph{International Conference on Machine Learning}, pages 1616--1630. PMLR, 2024.

\bibitem[Aumann(1976)]{aumann2016agreeing}
Robert~J. Aumann.
\newblock Agreeing to disagree.
\newblock \emph{Annals of Statistics}, 4\penalty0 (6):\penalty0 1236--1239, 1976.

\bibitem[Banerji et~al.(2023)Banerji, Chakraborti, Harbron, and MacArthur]{banerji2023clinical}
Christopher R.~S. Banerji, Tapabrata Chakraborti, Chris Harbron, and Ben~D. MacArthur.
\newblock Clinical {AI} tools must convey predictive uncertainty for each individual patient.
\newblock \emph{Nature Medicine}, 29\penalty0 (12):\penalty0 2996--2998, 2023.

\bibitem[Barber et~al.(2021)Barber, Candès, Ramdas, and Tibshirani]{barber2021limits}
Rina~Foygel Barber, Emmanuel~J. Candès, Aaditya Ramdas, and Ryan~J. Tibshirani.
\newblock The limits of distribution-free conditional predictive inference.
\newblock \emph{Information and Inference: A Journal of the IMA}, 10\penalty0 (2):\penalty0 455--482, 2021.

\bibitem[Barber et~al.(2023)Barber, Candès, Ramdas, and Tibshirani]{barber2023conformal}
Rina~Foygel Barber, Emmanuel~J. Candès, Aaditya Ramdas, and Ryan~J. Tibshirani.
\newblock Conformal prediction beyond exchangeability.
\newblock \emph{Annals of Statistics}, 51\penalty0 (2):\penalty0 816--845, 2023.

\bibitem[Belia et~al.(2005)Belia, Fidler, Williams, and Cumming]{belia2005researchers}
Sarah Belia, Fiona Fidler, Jennifer Williams, and Geoff Cumming.
\newblock Researchers misunderstand confidence intervals and standard error bars.
\newblock \emph{Psychological Methods}, 10\penalty0 (4):\penalty0 389--396, 2005.

\bibitem[Bengs et~al.(2022)Bengs, H{\"u}llermeier, and Waegeman]{bengs2022pitfalls}
Viktor Bengs, Eyke H{\"u}llermeier, and Willem Waegeman.
\newblock Pitfalls of epistemic uncertainty quantification through loss minimisation.
\newblock \emph{Advances in Neural Information Processing Systems}, 35:\penalty0 29205--29216, 2022.

\bibitem[Bengs et~al.(2023)Bengs, H{\"u}llermeier, and Waegeman]{bengs2023second}
Viktor Bengs, Eyke H{\"u}llermeier, and Willem Waegeman.
\newblock On second-order scoring rules for epistemic uncertainty quantification.
\newblock In \emph{International Conference on Machine Learning}, pages 2078--2091. PMLR, 2023.

\bibitem[Bhatnagar et~al.(2023)Bhatnagar, Wang, Xiong, and Bai]{bhatnagar2023improved}
Aadyot Bhatnagar, Huan Wang, Caiming Xiong, and Yu~Bai.
\newblock Improved online conformal prediction via strongly adaptive online learning.
\newblock In \emph{International Conference on Machine Learning}, pages 2337--2363. PMLR, 2023.

\bibitem[Bhatt et~al.(2021)Bhatt, Antor{\'a}n, Zhang, Liao, Sattigeri, Fogliato, Melan{\c{c}}on, Krishnan, Stanley, Tickoo, et~al.]{bhatt2021uncertainty}
Umang Bhatt, Javier Antor{\'a}n, Yunfeng Zhang, Q.~Vera Liao, Prasanna Sattigeri, Riccardo Fogliato, Gabrielle Melan{\c{c}}on, Ranganath Krishnan, Jason Stanley, Omesh Tickoo, et~al.
\newblock Uncertainty as a form of transparency: Measuring, communicating, and using uncertainty.
\newblock In \emph{Proceedings of the 2021 AAAI/ACM Conference on AI, Ethics, and Society}, pages 401--413, 2021.

\bibitem[Blasiok et~al.(2024)Blasiok, Gopalan, Hu, and Nakkiran]{blasiok2024does}
Jaroslaw Blasiok, Parikshit Gopalan, Lunjia Hu, and Preetum Nakkiran.
\newblock When does optimizing a proper loss yield calibration?
\newblock In \emph{Advances in Neural Information Processing Systems}, volume~36, 2024.

\bibitem[Collina et~al.(2024)Collina, Goel, Gupta, and Roth]{collina2024tractable}
Natalie Collina, Surbhi Goel, Varun Gupta, and Aaron Roth.
\newblock Tractable agreement protocols.
\newblock \emph{arXiv preprint arXiv:2411.19791}, 2024.

\bibitem[Cortes-Gomez et~al.(2024)Cortes-Gomez, Pati{\~n}o, Byun, Wu, Horvitz, and Wilder]{cortes2024decision}
Santiago Cortes-Gomez, Carlos Pati{\~n}o, Yewon Byun, Steven Wu, Eric Horvitz, and Bryan Wilder.
\newblock Decision-focused uncertainty quantification.
\newblock \emph{arXiv preprint arXiv:2410.01767}, 2024.

\bibitem[Corvelo~Benz and Rodriguez(2024)]{corvelo2024human}
Nina Corvelo~Benz and Manuel Rodriguez.
\newblock Human-aligned calibration for {AI}-assisted decision making.
\newblock In \emph{Advances in Neural Information Processing Systems}, volume~36, 2024.

\bibitem[Cresswell et~al.(2024{\natexlab{a}})Cresswell, Kumar, Sui, and Belbahri]{cresswell2024conformal}
Jesse~C. Cresswell, Bhargava Kumar, Yi~Sui, and Mouloud Belbahri.
\newblock Conformal prediction sets can cause disparate impact.
\newblock \emph{arXiv preprint arXiv:2410.01888}, 2024{\natexlab{a}}.

\bibitem[Cresswell et~al.(2024{\natexlab{b}})Cresswell, Sui, Kumar, and Vouitsis]{cresswell2024}
Jesse~C. Cresswell, Yi~Sui, Bhargava Kumar, and No{\"e}l Vouitsis.
\newblock Conformal prediction sets improve human decision making.
\newblock In \emph{International Conference on Machine Learning}, pages 9439--9457. PMLR, 2024{\natexlab{b}}.

\bibitem[Davison et~al.(2003)Davison, Hinkley, and Young]{davison2003recent}
Anthony~C. Davison, David~V. Hinkley, and G.~Alastair Young.
\newblock Recent developments in bootstrap methodology.
\newblock \emph{Statistical Science}, 18:\penalty0 141--157, 2003.

\bibitem[Dawid(1985)]{dawid1985calibration}
A.~Philip Dawid.
\newblock Calibration-based empirical probability.
\newblock \emph{Annals of Statistics}, 13\penalty0 (4):\penalty0 1251--1274, 1985.

\bibitem[Deng et~al.(2009)Deng, Dong, Socher, Li, Li, and Fei-Fei]{deng2009imagenet}
Jia Deng, Wei Dong, Richard Socher, Li-Jia Li, Kai Li, and Li~Fei-Fei.
\newblock Imagenet: A large-scale hierarchical image database.
\newblock In \emph{2009 IEEE Conference on Computer Vision and Pattern Recognition}, pages 248--255. Ieee, 2009.

\bibitem[Depeweg et~al.(2018)Depeweg, Hernandez-Lobato, Doshi-Velez, and Udluft]{depeweg2018decomposition}
Stefan Depeweg, Jose-Miguel Hernandez-Lobato, Finale Doshi-Velez, and Steffen Udluft.
\newblock Decomposition of uncertainty in {Bayesian} deep learning for efficient and risk-sensitive learning.
\newblock In \emph{International Conference on Machine Learning}, pages 1184--1193. PMLR, 2018.

\bibitem[Dwork et~al.(2021)Dwork, Kim, Reingold, Rothblum, and Yona]{dwork2021outcome}
Cynthia Dwork, Michael~P. Kim, Omer Reingold, Guy~N. Rothblum, and Gal Yona.
\newblock Outcome indistinguishability.
\newblock In \emph{Proceedings of the 53rd Annual ACM SIGACT Symposium on Theory of Computing}, pages 1095--1108, 2021.

\bibitem[Feldman et~al.(2023)Feldman, Ringel, Bates, and Romano]{feldman2023achieving}
Shai Feldman, Liran Ringel, Stephen Bates, and Yaniv Romano.
\newblock Achieving risk control in online learning settings.
\newblock \emph{Transactions on Machine Learning Research}, 2023.
\newblock ISSN 2835-8856.
\newblock URL \url{https://openreview.net/forum?id=5Y04GWvoJu}.

\bibitem[Foster and Vohra(1998)]{foster1998asymptotic}
Dean~P. Foster and Rakesh~V. Vohra.
\newblock Asymptotic calibration.
\newblock \emph{Biometrika}, 85\penalty0 (2):\penalty0 379--390, 1998.

\bibitem[Fudenberg et~al.(2017)Fudenberg, Romanyuk, and Strack]{fudenberg2017active}
Drew Fudenberg, Gleb Romanyuk, and Philipp Strack.
\newblock Active learning with a misspecified prior.
\newblock \emph{Theoretical Economics}, 12\penalty0 (3):\penalty0 1155--1189, 2017.

\bibitem[Gal and Ghahramani(2016)]{gal2016dropout}
Yarin Gal and Zoubin Ghahramani.
\newblock Dropout as a {Bayesian} approximation: Representing model uncertainty in deep learning.
\newblock In \emph{International Conference on Machine Learning}, pages 1050--1059. PMLR, 2016.

\bibitem[Garcia-Galindo et~al.(2023)Garcia-Galindo, Lopez-De-Castro, and Armananzas]{garcia2023uncertainty}
Alberto Garcia-Galindo, Marcos Lopez-De-Castro, and Rub{\'e}n Armananzas.
\newblock An uncertainty-aware sequential approach for predicting response to neoadjuvant therapy in breast cancer.
\newblock In \emph{Conformal and Probabilistic Prediction with Applications}, pages 74--88. PMLR, 2023.

\bibitem[Gelman(2011)]{gelman2011}
Andrew Gelman.
\newblock The pervasive twoishness of statistics; in particular, the ``sampling distribution'' and the ``likelihood'' are two different models, and that's a good thing.
\newblock \emph{Statistical Modeling, Causal Inference, and Social Science}, 2011.
\newblock URL \url{https://statmodeling.stat.columbia.edu/2011/06/20/the_sampling_di_1/}.

\bibitem[Gibbs and Candès(2021)]{gibbs2021adaptive}
Isaac Gibbs and Emmanuel~J. Candès.
\newblock Adaptive conformal inference under distribution shift.
\newblock In \emph{Advances in Neural Information Processing Systems}, volume~34, pages 1660--1672, 2021.

\bibitem[Gibbs and Candès(2024)]{gibbs2024conformal}
Isaac Gibbs and Emmanuel~J. Candès.
\newblock Conformal inference for online prediction with arbitrary distribution shifts.
\newblock \emph{Journal of Machine Learning Research}, 25\penalty0 (162):\penalty0 1--36, 2024.

\bibitem[Gibbs et~al.(2023)Gibbs, Cherian, and Candès]{gibbs2023conformal}
Isaac Gibbs, John~J Cherian, and Emmanuel~J. Candès.
\newblock Conformal prediction with conditional guarantees.
\newblock \emph{arXiv preprint arXiv:2305.12616}, 2023.

\bibitem[Gopalan et~al.(2021)Gopalan, Kalai, Reingold, Sharan, and Wieder]{gopalan2021omnipredictors}
Parikshit Gopalan, Adam~Tauman Kalai, Omer Reingold, Vatsal Sharan, and Udi Wieder.
\newblock Omnipredictors.
\newblock \emph{arXiv preprint arXiv:2109.05389}, 2021.

\bibitem[Gopalan et~al.(2022)Gopalan, Kim, Singhal, and Zhao]{gopalan2022low}
Parikshit Gopalan, Michael~P. Kim, Mihir~A. Singhal, and Shengjia Zhao.
\newblock Low-degree multicalibration.
\newblock In \emph{Conference on Learning Theory}, pages 3193--3234. PMLR, 2022.

\bibitem[Grove et~al.(2000)Grove, Zald, Lebow, Snitz, and Nelson]{grove2000clinical}
William~M. Grove, David~H. Zald, Boyd~S. Lebow, Beth~E. Snitz, and Chad Nelson.
\newblock Clinical versus mechanical prediction: A meta-analysis.
\newblock \emph{Psychological Assessment}, 12\penalty0 (1):\penalty0 19, 2000.

\bibitem[Guo et~al.(2017)Guo, Pleiss, Sun, and Weinberger]{guo2017calibration}
Chuan Guo, Geoff Pleiss, Yu~Sun, and Kilian~Q. Weinberger.
\newblock On calibration of modern neural networks.
\newblock In \emph{International Conference on Machine Learning}, pages 1321--1330. PMLR, 2017.

\bibitem[Guo et~al.(2024{\natexlab{a}})Guo, Wu, Hartline, and Hullman]{guo2024unexploited}
Ziyang Guo, Yifan Wu, Jason Hartline, and Jessica Hullman.
\newblock Unexploited information value in human-ai collaboration.
\newblock \emph{arXiv preprint arXiv:2411.10463}, 2024{\natexlab{a}}.

\bibitem[Guo et~al.(2024{\natexlab{b}})Guo, Wu, Hartline, and Hullman]{guo2024decision}
Ziyang Guo, Yifan Wu, Jason~D. Hartline, and Jessica Hullman.
\newblock A decision theoretic framework for measuring {AI} reliance.
\newblock In \emph{The 2024 ACM Conference on Fairness, Accountability, and Transparency}, pages 221--236, 2024{\natexlab{b}}.

\bibitem[H{\'e}bert-Johnson et~al.(2018)H{\'e}bert-Johnson, Kim, Reingold, and Rothblum]{hebert2018multicalibration}
Ursula H{\'e}bert-Johnson, Michael Kim, Omer Reingold, and Guy~N. Rothblum.
\newblock Multicalibration: Calibration for the (computationally-identifiable) masses.
\newblock In \emph{International Conference on Machine Learning}, pages 1939--1948. PMLR, 2018.

\bibitem[Heidhues et~al.(2018)Heidhues, K{\H{o}}szegi, and Strack]{heidhues2018unrealistic}
Paul Heidhues, Botond K{\H{o}}szegi, and Philipp Strack.
\newblock Unrealistic expectations and misguided learning.
\newblock \emph{Econometrica}, 86\penalty0 (4):\penalty0 1159--1214, 2018.

\bibitem[Hoekstra et~al.(2014)Hoekstra, Morey, Rouder, and Wagenmakers]{hoekstra2014robust}
Rink Hoekstra, Richard~D. Morey, Jeffrey~N Rouder, and Eric-Jan Wagenmakers.
\newblock Robust misinterpretation of confidence intervals.
\newblock \emph{Psychonomic Bulletin \& Review}, 21:\penalty0 1157--1164, 2014.

\bibitem[Holmes(2003)]{holmes2003bootstrapping}
Susan Holmes.
\newblock Bootstrapping phylogenetic trees: Theory and methods.
\newblock \emph{Statistical Science}, 18\penalty0 (2):\penalty0 241--255, 2003.

\bibitem[Hu and Wu(2024)]{hu2024predict}
Lunjia Hu and Yifan Wu.
\newblock Predict to minimize swap regret for all payoff-bounded tasks.
\newblock \emph{2024 IEEE 65th Annual Symposium on Foundations of Computer Ccience (FOCS)}, 2024.

\bibitem[Hullman and Gelman(2021)]{hullman2021designing}
Jessica Hullman and Andrew Gelman.
\newblock Designing for interactive exploratory data analysis requires theories of graphical inference.
\newblock \emph{Harvard Data Science Review}, 3\penalty0 (3):\penalty0 10--1162, 2021.

\bibitem[Hullman et~al.(2015)Hullman, Resnick, and Adar]{hullman2015hypothetical}
Jessica Hullman, Paul Resnick, and Eytan Adar.
\newblock Hypothetical outcome plots outperform error bars and violin plots for inferences about reliability of variable ordering.
\newblock \emph{PloS One}, 10\penalty0 (11):\penalty0 e0142444, 2015.

\bibitem[Hullman et~al.(2024)Hullman, Kale, and Hartline]{hullman2024decision}
Jessica Hullman, Alex Kale, and Jason Hartline.
\newblock Decision theoretic foundations for experiments evaluating human decisions.
\newblock \emph{arXiv preprint arXiv:2401.15106}, 2024.

\bibitem[Jung et~al.(2023)Jung, Noarov, Ramalingam, and Roth]{jung2023batch}
Christopher Jung, Georgy Noarov, Ramya Ramalingam, and Aaron Roth.
\newblock Batch multivalid conformal prediction.
\newblock In \emph{International Conference on Learning Representations}, 2023.
\newblock URL \url{https://openreview.net/forum?id=Dk7QQp8jHEo}.

\bibitem[Kale et~al.(2020)Kale, Kay, and Hullman]{kale2020visual}
Alex Kale, Matthew Kay, and Jessica Hullman.
\newblock Visual reasoning strategies for effect size judgments and decisions.
\newblock \emph{IEEE Transactions on Visualization and Computer Graphics}, 27\penalty0 (2):\penalty0 272--282, 2020.

\bibitem[Kay et~al.(2016)Kay, Kola, Hullman, and Munson]{kay2016ish}
Matthew Kay, Tara Kola, Jessica~R. Hullman, and Sean~A. Munson.
\newblock When (ish) is my bus? user-centered visualizations of uncertainty in everyday, mobile predictive systems.
\newblock In \emph{Proceedings of the 2016 CHI Conference on Human Factors in Computing Systems}, pages 5092--5103, 2016.

\bibitem[Kiyani et~al.(2025)Kiyani, Pappas, Roth, and Hassani]{kiyani2025decision}
Shayan Kiyani, George Pappas, Aaron Roth, and Hamed Hassani.
\newblock Decision theoretic foundations for conformal prediction: Optimal uncertainty quantification for risk-averse agents.
\newblock \emph{arXiv preprint arXiv:2502.02561}, 2025.

\bibitem[Kleinberg et~al.(2023)Kleinberg, Leme, Schneider, and Teng]{kleinberg2023u}
Bobby Kleinberg, Renato~Paes Leme, Jon Schneider, and Yifeng Teng.
\newblock U-calibration: Forecasting for an unknown agent.
\newblock In \emph{The Thirty Sixth Annual Conference on Learning Theory}, pages 5143--5145. PMLR, 2023.

\bibitem[Lu et~al.(2022)Lu, Lemay, Chang, Höbel, and Kalpathy-Cramer]{lu2022fair}
Charles Lu, Andréanne Lemay, Ken Chang, Katharina Höbel, and Jayashree Kalpathy-Cramer.
\newblock Fair conformal predictors for applications in medical imaging.
\newblock \emph{Proceedings of the AAAI Conference on Artificial Intelligence}, 36\penalty0 (11):\penalty0 12008--12016, Jun. 2022.
\newblock \doi{10.1609/aaai.v36i11.21459}.

\bibitem[Manski(2009)]{manski2009}
Charles~F Manski.
\newblock The 2009 {Lawrence R. Klein} lecture: Diversified treatment under ambiguity.
\newblock \emph{International Economic Review}, 50\penalty0 (4):\penalty0 1013--1041, 2009.

\bibitem[Mat{\v{e}}jka and McKay(2015)]{matvejka2015rational}
Filip Mat{\v{e}}jka and Alisdair McKay.
\newblock Rational inattention to discrete choices: A new foundation for the multinomial logit model.
\newblock \emph{American Economic Review}, 105\penalty0 (1):\penalty0 272--298, 2015.

\bibitem[Meehl(1954)]{meehl1954clinical}
Paul~E. Meehl.
\newblock \emph{Clinical Versus Statistical Prediction: A Theoretical Analysis and a Review of the Evidence}.
\newblock University of Minnesota Press, 1954.

\bibitem[Minderer et~al.(2021)Minderer, Djolonga, Romijnders, Hubis, Zhai, Houlsby, Tran, and Lucic]{minderer2021revisiting}
Matthias Minderer, Josip Djolonga, Rob Romijnders, Frances Hubis, Xiaohua Zhai, Neil Houlsby, Dustin Tran, and Mario Lucic.
\newblock Revisiting the calibration of modern neural networks.
\newblock \emph{Advances in Neural Information Processing Systems}, 34:\penalty0 15682--15694, 2021.

\bibitem[Morgan and Manning(1985)]{morgan1985optimal}
Peter Morgan and Richard Manning.
\newblock Optimal search.
\newblock \emph{Econometrica}, 53:\penalty0 923--944, 1985.

\bibitem[Neal(2012)]{neal2012bayesian}
Radford~M. Neal.
\newblock \emph{Bayesian Learning for Neural Networks}.
\newblock Springer, 2012.

\bibitem[Platt(1999)]{platt1999probabilistic}
John~C Platt.
\newblock Probabilistic outputs for support vector machines and comparisons to regularized likelihood methods.
\newblock \emph{Advances in Large Margin Classifiers}, 10\penalty0 (3):\penalty0 61--74, 1999.

\bibitem[Podkopaev and Ramdas(2021)]{podkopaev2021distribution}
Aleksandr Podkopaev and Aaditya Ramdas.
\newblock Distribution-free uncertainty quantification for classification under label shift.
\newblock In \emph{Uncertainty in Artificial Intelligence}, pages 844--853. PMLR, 2021.

\bibitem[Prinster et~al.(2024)Prinster, Stanton, Liu, and Saria]{prinster2024conformal}
Drew Prinster, Samuel~Don Stanton, Anqi Liu, and Suchi Saria.
\newblock Conformal validity guarantees exist for any data distribution (and how to find them).
\newblock In \emph{International Conference on Machine Learning}, pages 41086--41118. PMLR, 2024.

\bibitem[Romano et~al.(2020)Romano, Sesia, and Candès]{romano2020classification}
Yaniv Romano, Matteo Sesia, and Emmanuel~J. Candès.
\newblock Classification with valid and adaptive coverage.
\newblock In \emph{Advances in Neural Information Processing Systems}, volume~33, pages 3581--3591, 2020.

\bibitem[Savage(1972)]{savage1972foundations}
Leonard~J. Savage.
\newblock \emph{The Foundations of Statistics}.
\newblock Courier Corporation, 1972.

\bibitem[Sims(2003)]{sims2003implications}
Christopher~A Sims.
\newblock Implications of rational inattention.
\newblock \emph{Journal of monetary Economics}, 50\penalty0 (3):\penalty0 665--690, 2003.

\bibitem[Steyvers et~al.(2022)Steyvers, Tejeda, Kerrigan, and Smyth]{steyvers2022bayesian}
Mark Steyvers, Heliodoro Tejeda, Gavin Kerrigan, and Padhraic Smyth.
\newblock Bayesian modeling of human-{AI} complementarity.
\newblock \emph{Proceedings of the National Academy of Sciences}, 119\penalty0 (11):\penalty0 e2111547119, 2022.

\bibitem[Stigler(1961)]{stigler1961economics}
George~J. Stigler.
\newblock The economics of information.
\newblock \emph{Journal of Political Economy}, 69\penalty0 (3):\penalty0 213--225, 1961.

\bibitem[Straitouri and Rodriguez(2023)]{straitouri2023designing}
Eleni Straitouri and Manuel~Gomez Rodriguez.
\newblock Designing decision support systems using counterfactual prediction sets.
\newblock \emph{arXiv preprint arXiv:2306.03928}, 2023.

\bibitem[Straitouri et~al.(2023)Straitouri, Wang, Okati, and Rodriguez]{straitouri2023improving}
Eleni Straitouri, Lequn Wang, Nastaran Okati, and Manuel~Gomez Rodriguez.
\newblock Improving expert predictions with conformal prediction.
\newblock In \emph{International Conference on Machine Learning}, pages 32633--32653. PMLR, 2023.

\bibitem[Tibshirani et~al.(2019)Tibshirani, Foygel~Barber, Candès, and Ramdas]{tibshirani2019conformal}
Ryan~J. Tibshirani, Rina Foygel~Barber, Emmanuel~J. Candès, and Aaditya Ramdas.
\newblock Conformal prediction under covariate shift.
\newblock In \emph{Advances in Neural Information Processing Systems}, volume~32, 2019.

\bibitem[Vovk(2012)]{vovk2012conditional}
Vladimir Vovk.
\newblock Conditional validity of inductive conformal predictors.
\newblock In \emph{Asian Conference on Machine Learning}, pages 475--490. PMLR, 2012.

\bibitem[Vovk et~al.(2005)Vovk, Gammerman, and Shafer]{vovk2005algorithmic}
Vladimir Vovk, Alexander Gammerman, and Glenn Shafer.
\newblock \emph{Algorithmic Learning in a Random World}.
\newblock Springer, 2005.

\bibitem[Wald(1949)]{wald1949statistical}
Abraham Wald.
\newblock Statistical decision functions.
\newblock \emph{Annals of Mathematical Statistics}, 20:\penalty0 165--205, 1949.

\bibitem[Wilder et~al.(2020)Wilder, Horvitz, and Kamar]{wilder2020learning}
Bryan Wilder, Eric Horvitz, and Ece Kamar.
\newblock Learning to complement humans.
\newblock \emph{arXiv preprint arXiv:2005.00582}, 2020.

\bibitem[Wu et~al.(2023)Wu, Guo, Mamakos, Hartline, and Hullman]{wu2023rational}
Yifan Wu, Ziyang Guo, Michalis Mamakos, Jason Hartline, and Jessica Hullman.
\newblock The rational agent benchmark for data visualization.
\newblock \emph{IEEE Transactions on Visualization and Computer Graphics}, 2023.

\bibitem[Zhang et~al.(2024)Zhang, Chatzimparmpas, Kamali, and Hullman]{zhang2024}
Dongping Zhang, Angelos Chatzimparmpas, Negar Kamali, and Jessica Hullman.
\newblock Evaluating the utility of conformal prediction sets for ai-advised image labeling.
\newblock In \emph{Proceedings of the CHI Conference on Human Factors in Computing Systems}, pages 1--19, 2024.

\bibitem[Zhang and Maloney(2012)]{zhang2012ubiquitous}
Hang Zhang and Laurence~T Maloney.
\newblock Ubiquitous log odds: a common representation of probability and frequency distortion in perception, action, and cognition.
\newblock \emph{Frontiers in neuroscience}, 6:\penalty0 1, 2012.

\bibitem[Zhao et~al.(2021)Zhao, Kim, Sahoo, Ma, and Ermon]{zhao2021calibrating}
Shengjia Zhao, Michael Kim, Roshni Sahoo, Tengyu Ma, and Stefano Ermon.
\newblock Calibrating predictions to decisions: A novel approach to multi-class calibration.
\newblock In \emph{Advances in Neural Information Processing Systems}, volume~34, pages 22313--22324, 2021.

\end{thebibliography}

\end{document}